\documentclass{article} \usepackage{iclr2026_conference,times}

\usepackage{amsmath,amsfonts,bm}

\def\eqref#1{equation~\ref{#1}}

\def\1{\bm{1}}

\DeclareMathAlphabet{\mathsfit}{\encodingdefault}{\sfdefault}{m}{sl}
\SetMathAlphabet{\mathsfit}{bold}{\encodingdefault}{\sfdefault}{bx}{n}

 \usepackage{times}
\usepackage[utf8]{inputenc} \usepackage[T1]{fontenc}    \usepackage{url}            \usepackage{booktabs}       \usepackage{nicefrac}       \usepackage{microtype}      \usepackage{adjustbox}      \usepackage{graphics,color}
\usepackage{wrapfig}
\usepackage{algorithm}
\usepackage[noend]{algpseudocode}
\algrenewcommand\algorithmicindent{1em}
\algrenewcommand{\algorithmiccomment}[1]{\bgroup\hskip2em\textcolor{ourdarkgreen2}{//~\textsl{#1}}\egroup}

\usepackage{enumitem}

\usepackage{amsfonts}       \usepackage{amsmath}       \usepackage{amssymb}
\usepackage{xspace}
\usepackage{caption}
\usepackage{subcaption}

\usepackage{xr-hyper}

\usepackage{todonotes}
\usepackage{lineno}
\usepackage{comment}
\usepackage[breaklinks,colorlinks,allcolors=ourblue]{hyperref}
\usepackage[capitalise]{cleveref}

\makeatletter
\DeclareRobustCommand\onedot{\futurelet\@let@token\@onedot}
\def\@onedot{\ifx\@let@token.\else.\null\fi\xspace}
\makeatother

\newcommand{\eg}{e.g\onedot}

\makeatletter
\newcommand*{\addFileDependency}[1]{\typeout{(#1)}
  \@addtofilelist{#1}
  \IfFileExists{#1}{}{\typeout{No file #1.}}
}
\makeatother

 \def\eqref#1{(\ref{#1})}

\usepackage[dvipsnames]{xcolor}
\definecolor{ourblue}{rgb}{0.368,0.507,0.71}    \definecolor{ourorange}{rgb}{0.881,0.611,0.142} \definecolor{ourgreen}{rgb}{0.56,0.692,0.195}   \definecolor{ourred}{rgb}{0.923,0.386,0.209}    \definecolor{ourviolet}{rgb}{0.528,0.471,0.701} \definecolor{ourbrown}{rgb}{0.772,0.432,0.102}  \definecolor{ourazure}{rgb}{0.364,0.619,0.782}  \definecolor{ourolive}{rgb}{0.572,0.586,0.}     \definecolor{ourgray}{RGB}{102,88,84}           

\definecolor{ourblue2}{RGB}{9,134,223} \definecolor{ourdarkblue2}{RGB}{5,97,164} \definecolor{ourlightblue2}{RGB}{132,201,250} 

\definecolor{ourorange2}{RGB}{224,90,18} \definecolor{ourdarkorange2}{RGB}{160,63,9} \definecolor{ourlightorange2}{RGB}{246,175,137} 

\definecolor{ouryellow2}{RGB}{227,213,25} \definecolor{ourdarkyellow2}{RGB}{177,166,17} \definecolor{ourlightyellow2}{RGB}{242,235,140} 

\definecolor{ourpink2}{RGB}{247,24,139} \definecolor{ourdarkpink2}{RGB}{164,4,86} \definecolor{ourlightpink2}{RGB}{250,163,207} 

\definecolor{ourgreen2}{RGB}{159,198,52} \definecolor{ourdarkgreen2}{RGB}{109,138,30} \definecolor{ourlightgreen2}{RGB}{209,228,154} 

\definecolor{ourgray2}{RGB}{124,124,115} \definecolor{ourdarkgray2}{RGB}{87,87,81} \definecolor{ourlightgray2}{RGB}{194,194,189}

\newcommand{\myemph}[1]{\textbf{\color{ourblue} #1}}

\usepackage{listings}

\DeclareFixedFont{\ttb}{T1}{txtt}{bx}{n}{12} \DeclareFixedFont{\ttm}{T1}{txtt}{m}{n}{12}  

\usepackage{listings}

\definecolor{codegreen}{rgb}{0,0.6,0}
\definecolor{codegray}{rgb}{0.5,0.5,0.5}
\definecolor{codepurple}{rgb}{0.58,0,0.82}
\definecolor{backcolour}{rgb}{0.95,0.95,0.92}

                 \newcommand{\T}{\ensuremath{\top}}                
\newcommand{\mujoco}{MuJoCo\xspace}
\newcommand{\convex}{penalty-based\xspace}

\newcommand{\method}{DiffMJX\xspace}
\newcommand{\fbtrick}{straight-through-trick\xspace}

\newcommand{\p}{\ensuremath{p}}  
\newcommand{\M}{\ensuremath{M}}
\newcommand{\f}{\ensuremath{f}}
\newcommand{\J}{\ensuremath{J}}

\newcommand{\ak}{\ensuremath{a_{k}}}  
\newcommand{\xk}{\ensuremath{x_{k}}}  
\newcommand{\xkk}{\ensuremath{x_{k+1}}}
\newcommand{\dotv}{\ensuremath{\dot v}}

\newcommand{\aref}{\ensuremath{a_{\mathrm{ref}}}}
\newcommand{\fc}{\ensuremath{f_{\mathcal{C}}}}
\newcommand{\fe}{\ensuremath{\f_{\mathcal{E}}}}
\newcommand{\ff}{\ensuremath{\f_{\mathcal{F}}}}

\DeclareMathOperator*{\sg}{sg}
 \graphicspath{{graphics/}}

\usepackage{hyperref}
\usepackage{url}
\clearpage{}\usepackage{amsthm}

\newtheoremstyle{compactexample}{3pt}{3pt}{\normalfont}{0pt}{\itshape}{}{0.5em}{\thmname{#1}\thmnumber{ #2} -- \thmnote{#3:}}

\theoremstyle{compactexample}
\usepackage[framemethod=tikz]{mdframed}

\definecolor{exback}{RGB}{247,247,247} \definecolor{exframe}{RGB}{210,210,210} 

\surroundwithmdframed[
backgroundcolor=exback,
linecolor=exframe,
linewidth=0.3pt,
roundcorner=2pt,
skipabove=4pt,
skipbelow=4pt,
innerleftmargin=6pt,
innerrightmargin=6pt,
innertopmargin=4pt,
innerbottommargin=4pt,
]{example}

\surroundwithmdframed[
backgroundcolor=exback,
linecolor=exframe,
linewidth=0.3pt,
roundcorner=2pt,
skipabove=4pt,
skipbelow=4pt,
innerleftmargin=6pt,
innerrightmargin=6pt,
innertopmargin=6pt,
innerbottommargin=6pt,
]{example*}

\newtheorem*{example*}{Example}

\clearpage{}

\title{\centering Differentiable Simulation of Hard Contacts with Soft Gradients for Learning and Control}

\author{Anselm Paulus$^1$\thanks{Equal contribution.}\ \ ,
	A.\ Ren\'e Geist$^{1*}$,\\
    \textbf{Pierre Schumacher$^{2}$,
	V\'\i t Musil$^3$,
	Simon Rappenecker$^1$,}\\
	\textbf{Georg Martius$^1$}
	\\[3pt]
	$^1$ University of T\"ubingen,
	T\"ubingen, Germany\\
	$^2$ MyoLab.AI, New York City, USA\\
	$^3$ Masaryk University, Brno, Czechia
	\\[3pt]
	\texttt{\{anselm.paulus,rene.geist,georg.martius\}@uni-tuebingen.de}
}

\iclrfinalcopy 
\begin{document}

\maketitle

\begin{abstract}
Contact forces introduce discontinuities into robot dynamics that severely limit the use of simulators for gradient-based optimization.
Penalty-based simulators such as MuJoCo, soften contact resolution to enable gradient computation. However, realistically simulating hard contacts requires stiff solver settings, which leads to incorrect simulator gradients when using automatic differentiation. Contrarily, using non-stiff settings strongly increases the sim-to-real gap. We analyze penalty-based simulators to pinpoint why gradients degrade under hard contacts. Building on these insights, we propose DiffMJX, which couples adaptive time integration with penalty-based simulation to substantially improve gradient accuracy. A second challenge is that contact gradients vanish when bodies separate. To address this, we introduce contacts from distance (CFD) which combines penalty-based simulation with straight-through estimation. By applying CFD exclusively in the backward pass, we obtain informative pre-contact gradients while retaining physical realism.

Project page: \url{https://github.com/martius-lab/diffmjx}
\end{abstract}

\section{Introduction}\label{sec:intro}
Gradients have powered major advances in machine learning ranging from video language models to robot control. 
In robotics, imitation  and reinforcement learning widely rely on gradient-based optimization. Yet, despite the dominance of sim-to-real techniques that leverage robot simulators for policy learning ~\citep{Tan18:sim2real,leehutter2020:AnymalTerrain,andrychowicz_learning_2020,radosavovic_real-world_2024,li2022wasabi}, most methods conspicuously avoid using simulator gradients. This is a missed opportunity considering that simulator gradients offer a direct route to updating actions and learning model parameters. If these gradients were accurate, we could fit simulators to real data significantly narrowing the sim-to-real gap and accelerate policy learning -- enabling policies for new tasks to be trained  in seconds rather than hours. \emph{Given the utility of simulator gradients, what prevents their use in robot learning?}
In practice, two fundamental issues hinder the use of simulator gradients: (i) discontinuities arising from contacts yield \textbf{erroneous gradients}, and (ii), if objects do not touch, \textbf{contact gradients are zero}. In this work, we chart a path that tackles both challenges.

\vspace{-1.5mm}
\paragraph{Differentiating through contacts.}
The choice of contact model has significant implications on a simulator's differentiability. Complementary-based solvers, such as \citet{howell2022dojo}, compute contact forces exactly, such that sudden jumps in the dynamics aggravate gradient computation. Hence, recent literature on differentiating through complementary-based solvers proposes analytical reformulations of the dynamics, \eg based on the implicit function theorem \citep{werling2021nimble, howell2022dojo}, or resorts to randomized smoothing \citep{tassa2010stochastic, duchi2012randomized, suh2022bundled, bouyarmane2009potential, xu2010sampling}. While differing in  computational cost, both approaches have similar empirical performance \citep{pang2023quasidynamic, schwarke2024learning}.
To improve computational efficiency, \mujoco \citep{todorov2014mujococonvex} reformulates complementary constraints as a convex optimization problem, where a constraint's ability to generate force grows proportionally to the constraint violation.
While such a \convex simulator can be made smooth, gradient inaccuracies increase with the \emph{contact stiffness}, the \emph{relative contact velocity}, and the \emph{integration step size}. While common advice points to reducing the step size to improve gradient accuracy, using sufficiently small step sizes results in prohibitively slow simulation.

\vspace{-1.5mm}
\paragraph{Computing gradients of stiff simulations.} 
In the sequel, we show that gradient errors in \convex simulators arise from computing a discrete approximation of continuous physics -- put simply, from numerically integrating stiff differential equations. 
Consequently, the gradients obtained via automatic differentiation are “incorrect” insofar as they do not align with those of the underlying continuous system.
Although adaptive integration is well studied, it's rarely used in robotics and to the best of our knowledge its utility for differentiable simulation has not been explored. 
This perspective complements prior proposals for time-of-impact correction \citep{hu2020difftaichi, schwarke2024learning} in impulse-based simulators.
We incorporate \textbf{adaptive timestep integration} into MuJoCo XLA, incurring a slight computational overhead while obtaining correct gradients in the presence of hard contacts, and remaining compatible with existing MuJoCo libraries.

\begin{wrapfigure}[10]{R}{0.4\textwidth}
    \vspace{-1em}
    \centering
    \includegraphics[width=\linewidth]{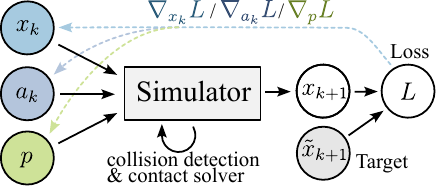}
    \caption{Common computational graph for robot control synthesis.}
    \label{fig:dynamics-autodiff-graph}
\end{wrapfigure}

\vspace{-1mm}
\paragraph{Computing gradients between non-colliding objects.} 
Another obstacle for gradient-based optimization for policy generation and system identification is the non-informativeness of gradients about unrealized contacts.
For example, when a robot's hand is not in contact with an object, then there is no gradient directing it to make contact for task facilitation. Therefore, drawing inspiration from prior works on contact-invariant optimization \citep{mordatch2012discovery, mordatch2012contact}, we propose \textbf{Contacts From Distance} (CFD) to address this challenge.
However, naively introducing artificial contact forces considerably alters the simulation, resulting in a too large sim-to-real gap.
In order to preserve the simulation realism, we propose using the \emph{\fbtrick} to introduce CFD solely in the gradient computation. 
We implement our changes in Mujoco XLA, and additionally fix some low-level collision routines to be truly differentiable. CFD complements position-based costs often used in robot learning by removing the need to specify exact contact locations.
\looseness=-1

In summary, DiffMJX and CFD provide additional tuning knobs to enable practitioners to set gradient correctness and inform an optimization on how non-colliding objects would need to move to influence each other's state. In turn, this work provides a new perspective on how to obtain useful gradients in penalty-based differentiable simulators thereby enabling parameter estimation and policy synthesis for collision-rich and high-dimensional systems.

\section{Robot simulation} \label{sec:mujoco}
\vspace{-2mm}
As illustrated in \cref{fig:dynamics-autodiff-graph}, we want to use automatic differentiation to obtain the \emph{correct} gradients
of a loss functional $L(\tilde x_{k+1}, \xk, \ak, p)$ where the next state of the robotic system is governed by the 
discrete-time dynamics $\xkk = \mathrm{step}(\xk, \ak, p)$ with the state $x_k:=x(t_k)=[q_k, v_k]$ at time $t_k$ consisting of the system's generalized position $q_k \in \mathbb{R}^{n_q}$ and velocity $v_k \in \mathbb{R}^{n_v}$, control actions $a_k\in \mathbb{R}^{n_a}$, and model parameters $\p\in\mathbb{R}^{n_p}$.
Typically, multi-body dynamics simulators consist of \emph{forward dynamics model} and a \emph{numerical integration method}.
The forward dynamics govern the system's acceleration $\dot v$ via the equations of motion
\begin{equation}
    \dotv = M^{-1} \left( \tau - c + \J^\T f\right) \label{eq:EOM}
\end{equation}
with the joint-space inertia matrix $\M(q) \in \mathbb{R}^{n_v \times n_v}$,
applied forces $\tau(x, a) \in \mathbb{R}^{n_v}$, bias force $c(x) \in \mathbb{R}^{n_v}$, constraint space Jacobian $J(q)\in\mathbb{R}^{n_c \times n_v}$,
and constraint forces $f(x) =\fe + \ff + \fc\in \mathbb{R}^{n_c}$ consisting of the equality constraint, the generalized friction, and the contact constraint forces.
The contact-free dynamics are typically derived via recursive multi-body algorithms \citep{featherstone2014rigidbody} while computing contacts are resolved through an intricate interplay of collision detection and contact force optimization. 
As we will see throughout this work, numerical integration plays a pivotal role in understanding how contact forces may hinder correct gradient computation.
For that, we will resort to \mujoco XLA as a concrete example of a \convex simulator.

\vspace{-4mm}
\subsection{Constraint resolution in MuJoCo XLA}
\vspace{-1mm}
MuJoCo XLA (MJX) is a reimplementation of \mujoco using the Python library JAX \citep{deepmind2020jax}, which enables GPU-parallelizable gradient computation via automatic differentiation.
\mujoco has become the de facto standard in robotics, alongside other widely used simulators such as Bullet \citep{coumans2021pybullet}, Drake \citep{tedrake2019drake}, DiffTaichi \citep{hu2020difftaichi} and Nvidia PhysX \citep{liang2018gpuacceleratedroboticsimulationdistributed}.
\mujoco’s importance is underpinned by NVIDIA and Google recently releasing \href{https://github.com/google-deepmind/mujoco_warp}{\mujoco Warp} \citep{mjwarp2025}.
In what follows, we provide a brief overview of how \mujoco resolves contacts, which serves as a hands-on example of a \convex simulator.

\vspace{-1mm}
\paragraph{Collision detection.}
Given the state $x$ and geometry parameterizations of two bodies, a collision detector returns the \emph{signed distance} $r$ between potential contact points alongside with their surface normals.  
A contact is only considered \emph{active} -- that is, a contact point can exert contact force -- if $r < 0$.
While collision detection is not the primary focus of this work, we later show how to remove discontinuities from its functions to ensure differentiability.

\begin{wrapfigure}[16]{R}{0.3\textwidth}
    \vspace{-1.5em}
    \centering
    \includegraphics[width=0.3\textwidth]{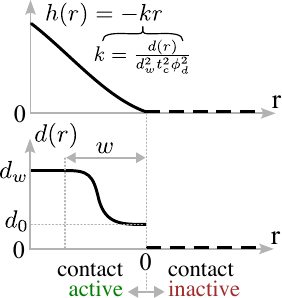}
    \vspace{-5mm}
    \caption{The position-level reference acceleration $h(r)$ and impedance $d(r)$ determine the contact force magnitudes that the solver can apply.}
    \label{fig:vanilla-mjx-solref}
\end{wrapfigure}

\vspace{-1mm}
\paragraph{Contact force solver.}
A detailed description of \mujoco is given in its documentation, here we provide a summary of key equations. 
Given \cref{eq:EOM}, \mujoco resolves constraints via a relaxation of Gauss' principle \citep{gauss1829neues} which in its primal formulation  reads
\begin{align}
\begin{split}
(\dot v, \dot \omega) &= \underset{(x,y)}{\arg\min} \bigl\| x - M^{-1}(\tau - c) \bigr\|_M^{2} + \|y - \aref\|_{R^{-1}}^{\mathrm{H}}
\\
\text{subject} \text{ to}&\:
J_{\mathcal{E}} x - y_e = 0,
\:
J_{\mathcal{F}} x - y_{\mathcal{F}} = 0,
\:
J_{\mathcal{C}}\, x - y_{\mathcal{C}} \in \mathcal{K}^*
\end{split} \label{eq:primal-problem}
\end{align}
with the friction cone (dual) $\mathcal{K}^*$, the regularizer $R>0$, and Huber norm $\| \cdot \|^{\mathrm{H}}$. The reference acceleration $\aref$ denotes the solver's target for the constraint space acceleration $\dot \omega$. Drawing inspiration from \cite{baumgarte1972stabilization}, $\aref$ follows a damped harmonic oscillator
\begin{equation}
    a_{\mathrm{ref},i} = -b_i (Jv)_i - k_i r_i = -\frac{2}{d_w \cdot t_c} (Jv)_i - \frac{d(r_i)}{d_w^2 \cdot t_c^2 \cdot \phi_d^2} r_i 
\end{equation}
whose dynamics are determined by the impedance $d(r)$, and the \emph{solref} parameters consisting of the time constant $t_c$ and the damping ratio $\phi_d$. The \emph{impedance} $d(r)$ is the central function for determining constraint forces (such as contact forces).
As illustrated in \cref{fig:vanilla-mjx-solref}, the impedance $d\in [0,1]$ is a function of the constraint violation $r$ and is specified by the \emph{solimp} parameters $(d_o, d_w, w, \text{midpoint}, \text{power})$ which define its shape as a polynomial spline. \mujoco's documentation refers to the \emph{impedance} as a ``constraints ability to generate force'' as it determines $\aref$ and weights the cost for applying constraint forces in \cref{eq:primal-problem} via the diagonal matrix $R$, which is computed as $R_{ii} = \hat A_{ii} (1-d_i)/d_i $,
where $\hat A$ denotes an approximation to $A=\text{diag}(J M^{-1} J^\T)$.
In turn, the constraint is hard if $R\rightarrow 0$, and approaches an infinitely soft (i.e. non-existent) constraint in the limit $R\rightarrow \infty$. If $\mathcal{K}^*$ solely contains pyramidal or elliptic cone constraints, then problem~\eqref{eq:primal-problem} reduces to a convex problem that is solved efficiently via an exact Newton method. \looseness=-1

\section{Correcting the contact gradients of \convex simulation}
\begin{figure}[t]
\vspace{-8mm}
\begin{minipage}{0.38\textwidth}
    \centering
    \vspace{-1mm}
    \includegraphics[width=\linewidth]{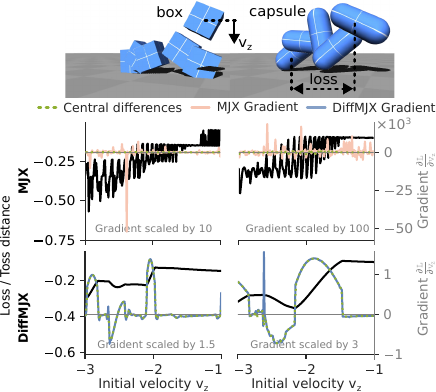}
    \caption{Simulation of primitives thrown onto a surface. For stiff contacts, MJX's gradients of the toss distance deviate from central differences, while \method maintains agreement.}
    \label{fig:object-tosses}
\end{minipage}
\hfill
\begin{minipage}{0.6\textwidth}
    \vspace{4mm}
    \centering
    \includegraphics[width=\textwidth]{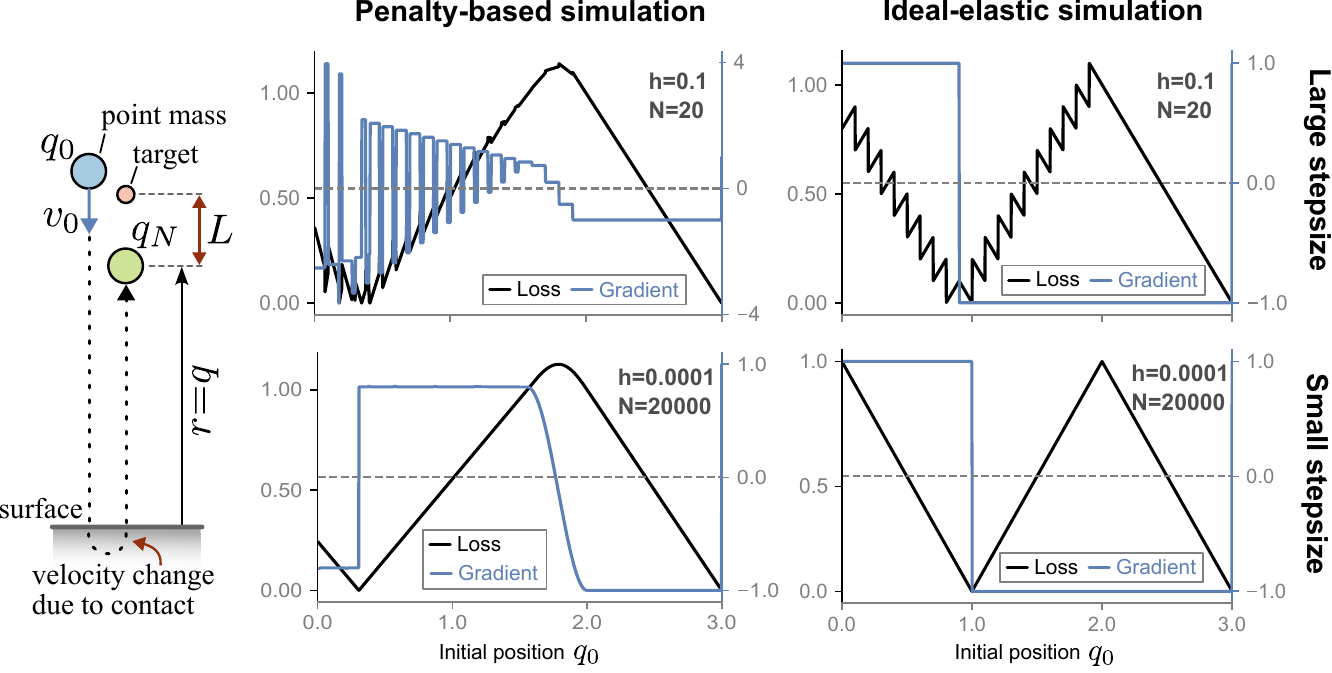}
\caption{Toy simulation of a point mass colliding with a surface that either resorts to an ideal-elastic contact model used in DiffTaichi or a \convex contact model similar to \mujoco. Decreasing the integration stepsize $h$ solely reduces errors in the penalty-based simulation, in the ideal-elastic simulation the gradient sign remains wrong.}
\label{fig:minimum-toy-problem}
\end{minipage}
\vspace{-5mm}
\end{figure}

To evaluate the correctness of gradients in the presence of contacts, as illustrated in the top row of \cref{fig:object-tosses}, we unroll the trajectory of several primitives bouncing against a plane to observe the final position and its gradient with respect to the initial velocity.
In \cref{fig:object-tosses} (MJX row), at the default integration step of 0.002\,s, we observe that the loss is oscillating at a high frequency, which results in large fluctuations of the gradient.
Similar oscillations have been discussed in previous works \citep{hu2020difftaichi,schwarke2024learning}. 
In particular, \cite{hu2020difftaichi} emphasizes that the ``time-of-impact'' (TOI) causes gradient oscillations when using ideal elastic or complementary collisions.
Yet, TOI gradient errors differ in nature from errors observed in \convex simulators; as illustrated in the example below.
We observe that state oscillations and the corresponding gradient artifacts are not a characteristic of the specific simulator.
Instead, more fundamentally, they arise from time-discretization errors in the ODE integration.
While this affects both penalty-based and ideal-elastic simulators, we will see that the issue has to be addressed differently.

\begin{example*}[Point collision] 
As illustrated in~\cref{fig:minimum-toy-problem} (right), a point mass starts at height $q_0$ with velocity $v_0=-1$ and is integrated for $N$ steps with semi-implicit Euler in the absence of gravity. It collides with a flat surface and bounces back up. Contact collision is either resolved via a minimal version of a penalty-based contact model as found in \mujoco or an ideal elastic collision as in \citep{hu2020difftaichi}. The corresponding JAX code is shown in \cref{fig:code-toi-toyexample} in the Appendix.
The loss $L=|q_N-q_T|$ is the distance between the point's final state $q_N$ and the target height $q_T=1$. For ideal-elastic collisions, we observe sawtooth-like loss oscillations in $L$, resulting in the gradient $\nabla_{q_0} L$ with \emph{the wrong sign}, independently of the stepsize as observed by \cite{hu2020difftaichi}. 
For the \convex simulation, gradient oscillations \emph{notably reduce when lowering the stepsize}.
\end{example*}

\paragraph{TOI correction does not fix gradients for \convex simulators, but small stepsizes do.} 
For an ideal elastic collision, the ODE of our minimal example is piecewise linear: The dynamics are linear (due to the absence of gravity) before and after the contact, at which the velocity is inverted (see also \cref{fig:code-toi-toyexample} in the Appendix).
The TOI approach proposed by \cite{hu2020difftaichi} exploits this structure by dynamically splitting the ODE into two linear segments at the time of contact. 
Integrating those separately thereby eliminates the discretization error and yields correct gradients.
In \convex simulation, the ODE is linear before and after the collision, but is non-linear with variable stiffness over the time of the collision.
Therefore, it cannot be easily divided into large linear segments.

Luckily, for penalty-based simulation we can instead rely on a different technique for reducing the gradient error.
The simple solution is to reduce the stepsize, which does not work in the ideal elastic case, as seen in \cref{fig:minimum-toy-problem}.
From an ODE perspective, this works because the penalty-based ODE is smooth, allowing us to continuously control the integration error by reducing the stepsize.
Unfortunately, simply reducing the step size is not a practical solution, as it necessitates extremely small steps that substantially increase the computational and memory demands of gradient computation. This trade-off raises a critical question: \emph{Can we retain correct gradients of realistic contacts while maintaining practical simulation speeds?}

\vspace{-2mm}
\subsection{Adaptive stepsize integration: Numerical precision on demand}

A standard method for integrating ODEs with variable stiffness is \emph{adaptive integration}. 
The idea behind adaptive integration is elegant: Two numerical integrators of different orders compute the next state. Their difference provides an estimate of the error. 
If the error is smaller than a given threshold, the step is accepted; otherwise, the step is rejected and the procedure is repeated with a different stepsize chosen by a feedback controller.
For further details on the rich history of adaptive stepsize integration, see \eg \cite{hairer2008solving-i,hairer2002solving-ii,soderlind2002automatic,soderlind2003digital}.

We use Diffrax \citep{kidger2021on} for efficient numerical integration in JAX, taking advantage of its solver flexibility and multiple backpropagation modes. 
\emph{Notably, as detailed in \cref{sec:appendix:diffrax}, we devoted substantial effort to seamlessly integrating quaternions and stateful actuators, which enables seamless compatibility between Diffrax and MJX while ensuring efficient adaptive integration and backwards compatibility with other \mujoco libraries.}

\begin{wrapfigure}[19]{r}{0.3\textwidth}
    \vspace{-2mm}
    \centering
    \includegraphics[width=0.3\textwidth]{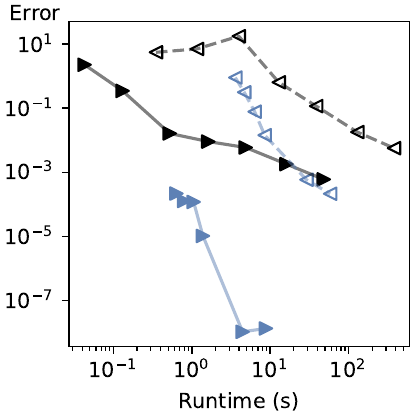}
\caption{
    Pareto front of gradient error vs forward runtime ($\blacktriangleright$) and gradient error vs backward runtime ($\triangleleft$) for standard semi-implicit Euler in MJX (\textbf{black}) and adaptive integration in \method{} ({\color{ourblue}\textbf{blue}}) on the cube bounce toy-example.
    }
    \label{fig:time-comparison-main}
\end{wrapfigure}

Testing our implementation on the cube bounce toy example, we observe in \cref{fig:time-comparison-main} that adaptive integration reduces the error in the loss and gradient by multiple orders of magnitude given the same computational budget; more detailed analysis is in \cref{fig:object-tosses-scatter} in the Appendix.
Importantly, the adaptive stepsize selection in Diffrax is independent across parallelized environments. Therefore, when vmapping across environments, contacts in one simulation do not slow down other simulations. For experimental verification, see \cref{fig:vmap_ball_drop} in
\cref{sec:app-sharp-bits}.

\paragraph{Resolving problems of Collision Detection.}

Using an adaptive integrator with MJX eliminates oscillations in the bounce example. However, gradients for some object primitives (capsule, cylinder, box) experience gradient artifacts due to non-differentiable operations in the collision detector arising from discrete case distinctions. 
We smoothed them with standard proxies, finally leading to the results in the bottom row of \cref{fig:object-tosses}, where analytical gradients nearly match central differences. Henceforth, we refer to MJX with the Diffrax integrator and smoothed collision detection as \method.

\begin{figure}[t]
\vspace{-8mm}
\begin{minipage}{0.3\textwidth}
    \includegraphics[width=\linewidth]{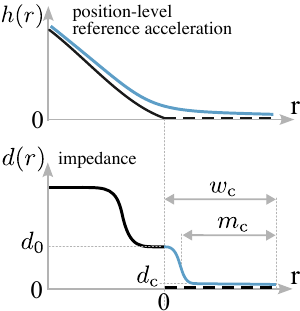}\\
    \vspace{-1mm}
    \caption{Contacts from distance (CFD): To let \mujoco create small contact forces between non-colliding objects, reference acceleration $h(r)$ and impedance $d(r)$ are adjusted to be nonzero for positive signed distances $r>0$.}
    \label{fig:cfg_impedance}
\end{minipage}
\hspace{1.5mm}
\begin{minipage}{0.35\textwidth}
    \centering
    \includegraphics[width=0.55\textwidth]{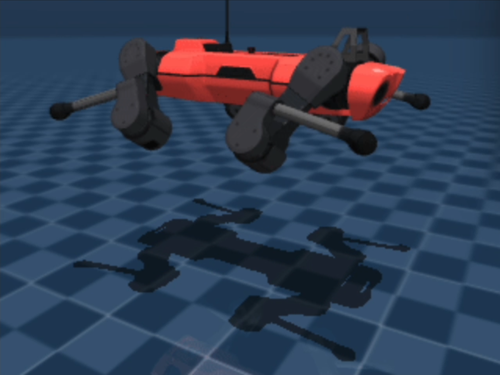}\\
    \vspace{1mm}
    \includegraphics[width=\linewidth]{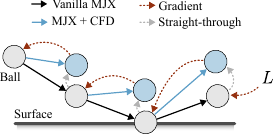}
    \caption{
    \myemph{Top:} Applying contact forces for $r>0$ in the forward pass of the simulation causes a robot to hover.
    \myemph{Bottom:} The \fbtrick is used to replace the original MJX derivative with the derivative of MJX + CFD, evaluated at the unaltered trajectory.}
    \label{fig:autodiff-forward-backward-trick-hovering}
\end{minipage}
\hspace{1.5mm}
\begin{minipage}{0.3\textwidth}
    \centering
    \vspace{-0.5mm}
    \includegraphics[width=\linewidth]{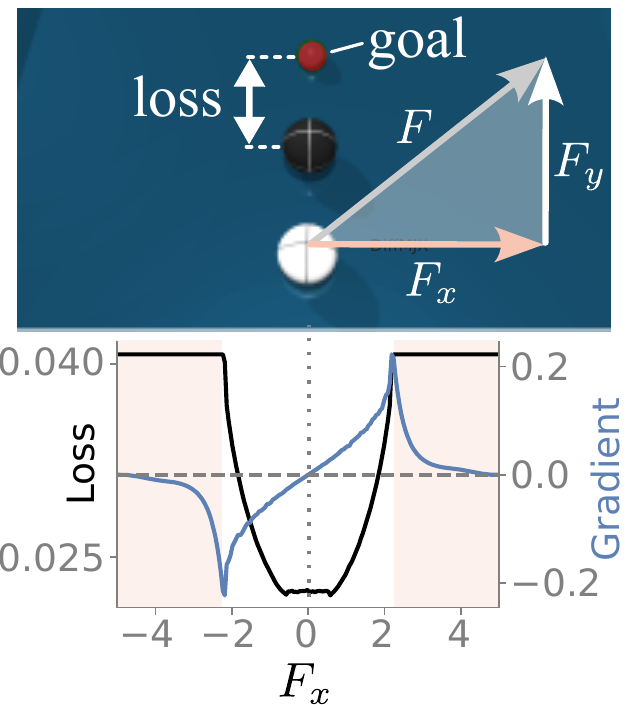}
\caption{Billiard simulation.
    \myemph{Top:} Force $F$ acts on the white ball affecting the loss.
    \myemph{Bottom:} Despite the loss derivative being zero if the balls do not collide, \method{} with CFD provides informative non-zero gradients.}
    \label{fig:billiard_loss}
\end{minipage}
\vspace{-4mm}
\end{figure}

\section{Contacts from distance with straight-through estimation}
While adaptive integration improves gradient accuracy, we now shift attention to computing 
\emph{informative gradients} between objects that are not in contact. To illustrate why the computation of such gradients is of fundamental importance for robot learning consider the following example.

\begin{example*}[Billiard shot] 
A billiard table is set up as shown in \cref{fig:billiard_loss} (top). 
At the first timestep, a force $F$ is exerted on the white ball such that it may hit the black ball. The optimization objective is the distance $L$ between the black ball and the target position. If the balls collide, MJX with adaptive integration yields non-zero gradients $\nabla_F L$. 
However, if $F$ does not cause the balls to touch, then $\nabla_F L$ is zero and therefore uninformative for optimization. 
\end{example*}

In the following, we propose \emph{contacts from distance} (CFD), a method for computing contact forces for positive signed distances $r$ in the gradient computation of a \convex simulation that yields informative gradients even if objects are not in contact. This is accomplished by \textbf{\emph{(i)} applying artificial contact forces} between non-colliding objects, and \textbf{\emph{(ii)} using artificial forces only in the gradient computation} to maintain simulation realism.

\newpage
\vspace*{-10mm}
\paragraph{Creating artificial contact forces.}
How to generate artificial contact forces in a \convex simulator depends on the respective contact model. Below, as a concrete example,  we propose a method tailored to \mujoco.
As discussed in \cref{sec:mujoco}, the magnitude of contact forces is determined by the impedance $d(r)$ and position-level reference acceleration $h(r)$. To enable the solver to apply CFD, $d(r)$ is augmented as shown in \cref{fig:cfg_impedance}.
Here, $d(r)$ remains unaltered for $r<0$ and is extended by an additional spline for $r>0$.
This continuation is parametrized by \emph{solimp-CFD} parameters $(d_c,d_0,w_c,m_c,p_c)$.
By default, the curve smoothly continues \mujoco's impedance at $d_0$ and tapers off to $d_c=0$ to ensure smooth differentiability.
The CFD-width $w_c$ specifies the distance for which artificial contact forces are generated. Moreover, we soften the reference acceleration $h(r)$ by replacing the ReLU function on the signed distance with a softplus (\cref{fig:cfg_impedance}).
This yields modified contact forces $f_\text{CFD}$.\looseness-1

\paragraph{Designing a surrogate gradient estimator.}
Naively adding CFD to a simulation produces nonphysical behaviors. As shown in \cref{fig:autodiff-forward-backward-trick-hovering} (top), the artificial contact forces would cause a quadruped to hover above the ground as if a soft foam mat of thickness $w_c$ had been placed on the surface.
As significantly altering simulation realism is not an option, we are faced with the question:
\emph{Can CFD be used to obtain informative contact gradients without affecting simulation realism?}

To positively answer this question,
we resort to the \emph{\fbtrick} on the ODE level:
\begin{align}
\dot{x}(t) &= \sg(F_\theta(t,x(t))) + \tilde{F}_\theta(t,x(t)) - \sg(\tilde{F}_\theta(t,x(t))),
\end{align}
where $\sg$ is the stop-gradient operator of an automatic differentiation library.
Here $F$ denotes the original ODE obtained from MJX or DiffMJX and $\tilde{F}$ denotes the ODE using CFD via 
\(
\dot v_\text{CFD} = M^{-1} \left( \tau - c + \J^\T f_\text{CFD}\right)
\).
In JAX this reads as:
\begin{lstlisting}
from jax.lax import stop_gradient
def forward(m: Model, d: Data) -> Data:
    d_mjx = _forward(m, d, cfd=False)  # Compute system acceleration without CFD
    d_cfd = _forward(m, d, cfd=True)   # Compute system acceleration using CFD
    grad_replace_fn = lambda x_mjx, x_cfd: stop_gradient(x_mjx) + x_cfd - stop_gradient(x_cfd)
    return jax.tree.map(grad_replace_fn, d_mjx, d_cfd)  # Reroute gradient computation
\end{lstlisting}

As illustrated in \cref{fig:autodiff-forward-backward-trick-hovering} (bottom), the above code ensures that the forward pass uses $F_\theta(t,x(t))$, whereas the backward pass deploys $\frac{\partial \tilde{F}_\theta}{\partial(x,\theta)}(t,x(t))$.
Crucially, the derivatives are evaluated at the unmodified forward trajectory $x(t)$.
While our approach requires the simulator's forward pass to be computed twice, the gradient is only evaluated once. As the gradient computation dominates the computational cost, CFD forms a practical method for improving contact gradients. 
Revisiting the billiard toss example (\cref{fig:billiard_loss}), 
using CFD with the \fbtrick yields informative gradients while keeping the loss unaltered.

Note that the \fbtrick is inspired by prior work on straight-through estimators~\citep{bengio2013estimating}; see also the related work in \cref{sec:related-work}. DiffMJX and CFD introduce easy-to-use tuning knobs in MJX to control gradient quality. \cref{sec:appendix:tuning-diffmjx} explains how to tune these knobs. In \cref{sec:appendix:diffrax}, we evaluate the \fbtrick with different autodiff techniques, including “Discretize-then-optimize” and “Optimize-then-discretize.”

\section{Evaluation}
In the following, we demonstrate the use of DiffMJX and CFD for learning physics parameters and computing robot control actions using gradient descent with simulator gradients.

\subsection{Parameter Identification} \label{sec:sysid}
Parameter identification in the presence of hard contacts remains a laborious task.
If contacts are hard, learning the dynamics of a cube requires impractical amounts of data for ``naive'' neural network regression \citep{parmar2021fundamental}. In comparison, \convex simulators can capture hard contacts, but the lack of correct gradients hinders efficient parameter estimation \citep{acosta2022validating}.
Therefore, recent work introduced intricate analytical pipelines for cube geometry estimation \citep{pfrommer2021contactnets, bianchini2023simultaneous} and graph-based networks for learning contact dynamics \citep{allen2023graphcontacts}. 
In what follows, we use the same real-world data as used in \cite{pfrommer2021contactnets, bianchini2023simultaneous, allen2023graphcontacts}. We demonstrate that \method{} with CDF enables simulator parameter estimation via standard gradient-based optimization.

\begin{figure}[t]
\vspace{-8mm}
\begin{minipage}{0.55\textwidth}
    \centering
    \includegraphics[width=\linewidth]{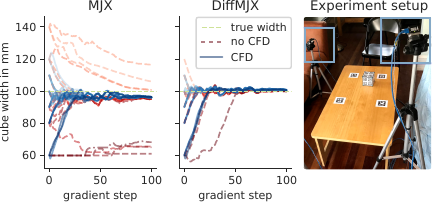}
    \vspace{-6mm}
    \caption{\myemph{Left:} Estimation of a cube's side length in MJX via gradient descent using multi-step ahead predictions.  \myemph{Right:} Experimental setup for collecting cube toss data. Image adapted from \cite{pfrommer2021contactnets}.}
    \label{fig:contactnet-sysid}
\end{minipage}
\hfill
\begin{minipage}{0.43\textwidth}
    \centering
    \vspace{4mm}
    \includegraphics[width=\linewidth]{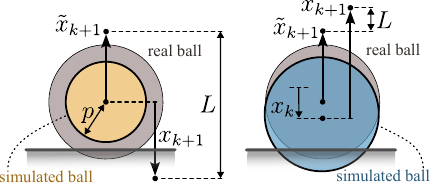}
    \vspace{-3mm}
    \caption{\myemph{Left:}~If the initial ball radius $p$ is set too small, then the gradient $\nabla_{\!p} L$ resulting from a one-step-ahead prediction is zero. \myemph{Right:} Large initial constraint penetration results in large gradients $\nabla_{\!x_k} L$.}
    \label{fig:estimation-problems}
\end{minipage}
\end{figure}

\vspace{-1mm}
\paragraph{Dataset and training setup.}
We use the Contactnets dataset \citep{pfrommer2021contactnets} which consists of $550$ trajectories of a 10\,cm acrylic cube that has been repeatedly tossed onto a wooden table. For training, trajectories are split into segments of length five such that the simulator is tasked to unroll four future steps starting from the initial state. Each segment and its prediction are fed to an $L_2$ loss whose gradient is used for gradient-based optimization using Adam \citep{kingma2015adam}. For systems with stiff dynamics, we favor multi-step-ahead predictions over one-step-ahead predictions, as they capture the cumulative effects of prediction errors over time. This setup enables a fairer analysis of MJX without CFD, as even for too small side length estimates as illustrated in \cref{fig:estimation-problems}, future state predictions can make contact to inform the optimization.

\begin{figure}[h]
\vspace{-1mm}
\begin{minipage}{\textwidth}
\centering
\includegraphics[width=0.45\linewidth, trim={0 0 34cm 0}, clip]{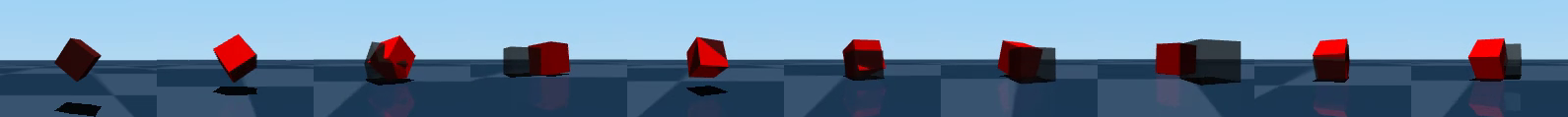}
\includegraphics[width=0.45\linewidth, trim={0 0 34cm 0}, clip]{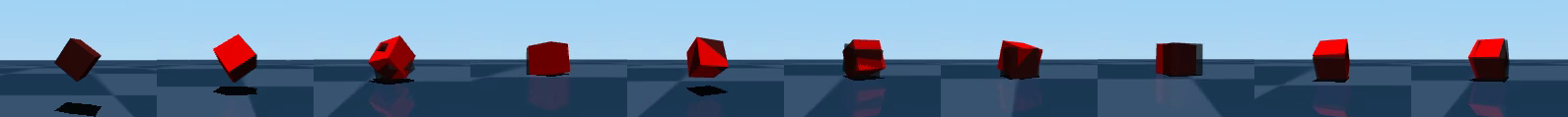}
\newline
\includegraphics[width=0.45\linewidth, trim={23cm 0 11cm 0}, clip]{graphics/output_bad.png}
\includegraphics[width=0.45\linewidth, trim={23cm 0 11cm 0}, clip]{graphics/output_good.png}
\newline
{\footnotesize Before training}
\hspace{5cm}
{\footnotesize After training}
\vspace{-1mm}
\caption{Comparison between real-world cube tosses ({\color{BrickRed}\textbf{red}}) and DiffMJX cube simulations ({\color{ourdarkgray2}\textbf{black}}).}
    \label{fig:cube_rollouts}
\end{minipage}
\vspace{-2mm}
\end{figure}

\vspace{-1mm}
\paragraph{Training results.}
The training results are shown in \cref{fig:contactnet-sysid}. Our version of MJX with smooth collision detection and MJX with CFD both achieve good estimation results with an error of around 5\% relative to the ground truth. If the side length is initialized at 60\,mm or 140\,mm, training either stalls fully or convergence is severely limited for MJX. The incorporation of CFD into MJX addresses convergence issues arising from poor initial parameters, while the integration of adaptive integration via DiffMJX significantly enhances estimation accuracy. DiffMJX improves estimation accuracy by dynamically adjusting the time steps during collisions, thereby mitigating time discretization errors. Further details and additional experiments in which also the contact parameters were identified are provided in \ref{sec:appendix-sysid}. \cref{fig:cube_rollouts} provides a comparison of DiffMJX's predictions after \emph{estimating contact and geometry parameters}.
To the best of our knowledge, we are the first to demonstrate parameter estimation of real-world cube dynamics using an automatically differentiable \convex simulator. While this represents a promising step forward, further experimentation is necessary to fully characterize the scope and limitations of this approach.

\subsection{Model Predictive Control}
Next, we conduct experiments on gradient-based model-predictive control.
At every plan step in the MPC loop, we refine a sequence of controls over a $256$-step horizon.
In the gradient-based planner, we compute gradients by backpropagating the differentiable cost computed on the rollout of the current plan through the MJX simulator.
The plan is then iteratively optimized using the Adam optimizer with a learning rate of $0.01$ for $32$ iterations.
Finally, the resulting plan is executed for $16$ steps in simulation, after which the planning procedure is repeated with the previous plan as a warm start.
As a baseline, we include a version of the predictive sampling planner from Mujoco MPC \citep{howell2022mjpc}, which at every plan step samples $k=\{64,256,1024\}$ trajectories, and executes the lowest-cost plan. For enabling a fair comparison, \emph{we significantly improved the performance of this planner for muscular systems by resorting to brown noise for sampling} \citep{pinneri2020icem}.

\begin{figure}
\centering
\begin{minipage}[t]{0.42\textwidth} \ \\[-0.5em]
\includegraphics[width=\linewidth]{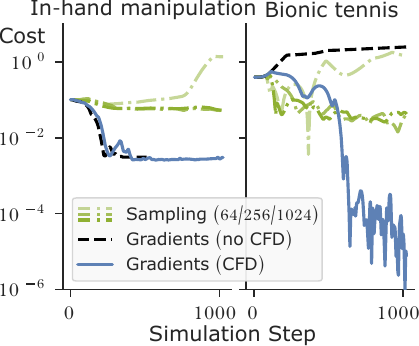}
\end{minipage}
\hfill
\begin{minipage}[t]{0.53\textwidth}\ \\[-1em]
\includegraphics[trim={0 0 0 5mm},clip, width=\linewidth]{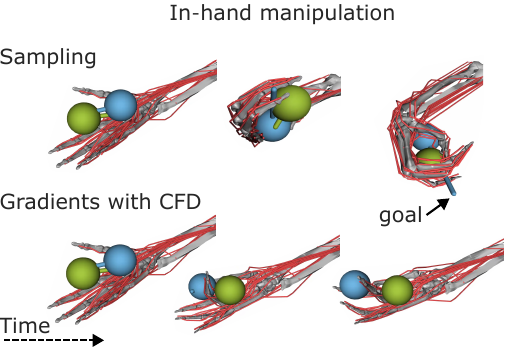}
\end{minipage}
\vspace{0mm}
\caption{
Autodiff-driven MPC with and without CFD on the in-hand manipulation and bionic tennis tasks (\cref{fig:ball-passing}). \textbf{In both tasks, only the distance between the ball and the respective goal is used as cost.}
\myemph{Left:}
Simulation cost evolution of gradient-based MPC, with and without contacts from distance (CFD), vs sampling-based MPC. The number for sampling indicates the number of samples used per planning step.
Sampling has difficulties solving the dexterous in-hand manipulation task; gradients without CFD cannot solve the bionic tennis task.
\myemph{Right:} 
Rendering of sampling-based MPC (1024 samples) vs gradient-based MPC with CFD on the in-hand manipulation task. The goal is to swap the balls, with the cost computed as L2 distance of the ball centers to positions fixed in the frame of the hand. The MyoHand model is actuated by 39 muscle-tendon units.
}
\label{fig:mpc-cost}
\vspace{-2mm}
\end{figure}

\begin{figure}[t]
    \centering
\includegraphics[width=\linewidth, trim={0 0 0 0}, clip]{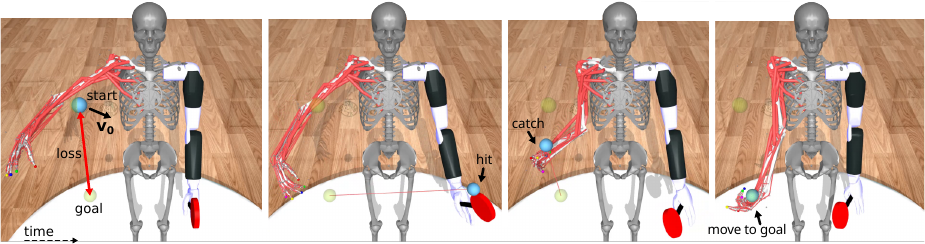}
    \caption{
Autodiff-driven MPC with CFD on the bionic tennis task. 
Task completion requires the racket to deflect the ball towards the MyoArm with 63 muscle-tendon actuators, which then catches the ball and moves it to the goal position. \textbf{Only the distance between ball and goal is used as cost.}
    }
    \label{fig:ball-passing}
\end{figure}

\paragraph{Models.}
As physical systems, we resort to state-of-the-art muscle-tendon models provided by MyoSuite \citep{caggiano2022myosuite,wang2022myosim}.
Models include the MyoHand (\cref{fig:mpc-cost}, right) adapted from the \href{https://sites.google.com/view/myochallenge}{MyoChallenge 2022}, which is comprised of 29 bones, 23 joints, and 39 muscle-tendon units.
We also use a bionic model (\cref{fig:ball-passing}) modified from the \href{https://sites.google.com/view/myosuite/myochallenge/myochallenge-2024}{MyoChallenge 2024}, which is comprised of the MyoArm with 27 degrees of freedom and 63 muscle-tendon units, and the simulated modular prosthetic limb with 26 degrees of freedom and 17 motor control units.

\paragraph{Dexterous in-hand manipulation.}

First, we consider an in-hand manipulation task, where the goal is to swap two balls in the MyoHand. 
The cost is given by the Euclidean distance between each of the balls and the respective target location, fixed in the frame of the hand.
Note that the muscle actuator implementation caused gradient errors, which we corrected using smooth functions as surrogates.
The results are reported in \cref{fig:mpc-cost}.
We find that gradient-based MPC can reliably solve this task, in contrast to the sampling-based planner.
Overparameterization in the muscle-tendon model with at least two muscles per joint actually benefits the gradient-based planner by helping escape local minima, similar to its role in optimizing overparametrized neural networks. 
In contrast, RL and sampling-based planners struggle with scaling in overparametrized higher-dimensional systems \citep{schumacher2023:deprl}. 
Thus, first-order methods using differentiable simulation should be able to tackle more complex control problems.
\looseness=-1

Notably, this task does not require contacts from distance because hand-ball interactions are frequent due to gravity.
Moreover, we identify two crucial components of the gradient-based MPC loop:
First is gradient clipping, which is important as the scale of gradients changes massively in the presence of contacts, as illustrated in \cref{fig:estimation-problems} (right).
This technique has also been reported to be effective in previous works on differentiable simulation \citep{xu2022shac,georgiev2024ahac}.
Second, we store the rollout cost of all gradient iterations and select the one with minimal cost. This is important as the cost landscape is highly non-convex, which is reflected in the non-monotonic cost evolution between the iterations of a planning step.

\paragraph{Bionic tennis: Using CFD to solve complex control tasks with minimal task supervision.}

Next, we test a more complex custom tennis task on the bionic model.
The task is to move a ball that is initially moving sideways to a target location below.
This can be achieved by bouncing it back using a racket that is welded to the prosthetic hand, and then catching it at the target location with the muscle hand.
In this task, \emph{the only cost supervision is again the Euclidean distance of the ball to the target}, the complicated sequential movement has to be discovered purely from this signal.

We report our findings in \cref{fig:mpc-cost} (left), see \cref{fig:ball-passing} for a rendering.
By design, the task initialization is such that the ball misses both hands, hence this task is not solvable by purely gradient-based MPC using vanilla MJX.
On the other hand, we observe that adding the CFD mechanism allows solving this task.
The sampling-based planner is a strong baseline in this task and gets close to solving it.
Initially, bouncing the ball back to the target only requires controlling the prosthetic arm, which is relatively low-dimensional. 
Hence, the sampling-based planner achieves this part easily.
However, as seen in the in-hand manipulation task, it struggles with precise control of the high-dimensional MyoHand, leading to sub-optimal results in balancing the ball at the goal position.

\begin{figure}[t]
\centering
\includegraphics[width=\linewidth]{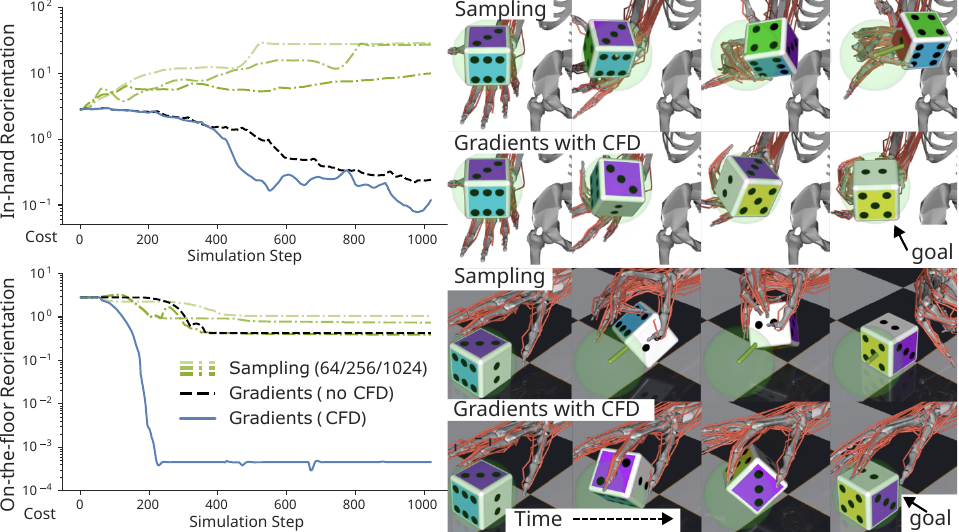}
\caption{
Autodiff-driven MPC with and without CFD on the in-hand and on-the-floor cube reorientation tasks. In both tasks, the loss consists of the L2 distance between the cube center and the target location, as well as a squared loss on the difference of quaternions describing the cube and target orientation.
\myemph{Top:}
Simulation cost evolution of gradient-based MPC, with and without contacts from distance (CFD), vs sampling-based MPC. The number for sampling indicates the number of samples used per planning step.
Sampling has difficulties solving the the in-hand reorientation task; only gradients with CFD can solve the on-the-table reorientation task perfectly.
\myemph{Right:} 
Rendering of sampling-based MPC (1024 samples) vs gradient-based MPC with CFD on the in-hand and on-the-table reorientation task. The goal is to reorient the cubes while staying at the target location (green sphere). The MyoHand model is actuated by 39 muscle-tendon units.
}
\label{fig:mpc-reorientation}
\vspace{-2mm}
\end{figure}

\newpage
\paragraph{Cube reorientation.}
Finally, we tackle another manipulation task of reorienting a cube to a desired target orientation, while staying at a target position. We include both an in-hand version and an on-the-table version of this task.

We report our findings in \cref{fig:mpc-reorientation}.
We observe that in the in-hand reorientation task, the sampling-based planner fails, while the gradient-based approach solves it. Similar to the ball-swap task, CFD does not make a big difference here, as contacts are automatically discovered from gravity.
In the on-the-table reorientation task, only the gradient-based planner with CFD activated solves the task perfectly. When deactivating CFD or using sampling, the planner still manages to reorient the cube to a satisfactory level, but fails to keep it in the target location.
Here, gravity alone does not result in all fingers making contact with the cuber, therefore CFD allows higher-fidelity control in comparison to CFD disabled.

\section{Conclusion}
In this work, we tackle two standing challenges of computing gradients of penalty-based simulations with hard contacts, namely gradient errors due to time discretization and zero contact gradients between non-colliding objects. A 1D~toy example illustrates how gradient errors are a consequence of time discretization and can be mitigated by reducing the integration stepsize. While reducing the stepsize improves gradient accuracy, it comes at the cost of increased simulation time and GPU memory usage.
To address these limitations, we introduce DiffMJX, a JAX library that incorporates the adaptive integration library Diffrax into \mujoco XLA. The use of adaptive time integration enables the reduction of discretization errors while significantly reducing memory overhead through checkpointing. 
Complementing DiffMJX, we propose Contacts from Distance (CFD), an extension of \convex contact resolution that improves gradient utility by introducing small virtual contact forces between near-interacting bodies, without affecting the forward simulation due to the \fbtrick.
We validated our contributions in a real-world system identification study, where DiffMJX accurately recovered a cube’s geometric parameters via gradient-based optimization. Beyond this, we present the first evidence that automatic differentiation through CFD can outperform sampling-based predictive planning on musculoskeletal manipulation tasks. We hope these results will guide future work on mitigating discretization errors in differentiable simulators and establish DiffMJX with CFD as a helpful toolbox for gradient-based robot learning.

\section*{Acknowledgements}
The authors thank Onur Beker for helpful discussions on softening the collision detection of MJX.
Addtionally, the authors thank Thomas Rupf, and Nico Gürtler for providing valuable feedback.

We thank the International Max Planck Research School for Intelligent Systems (IMPRS-IS) for supporting Anselm Paulus. 

This work was supported by the ERC - 101045454 REAL-RL and the German Federal Ministry of Education and Research (BMBF) through the Tübingen AI Center (FKZ: 01IS18039B). Georg Martius is a member of the Machine Learning Cluster of Excellence, EXC number 2064/1 – Project number 390727645.

The work on this paper was supported by the Czech Science Foundation (GAČR) grant no.~26-23981S.

\newpage
\bibliography{updated}
\bibliographystyle{icml2026_titleurl}

\newpage
\appendix

\section{Related Work} \label{sec:related-work}
\begin{wrapfigure}[21]{r}{0.35\textwidth}
\vspace{-2em}
    \centering
    \includegraphics[width=0.35\textwidth]{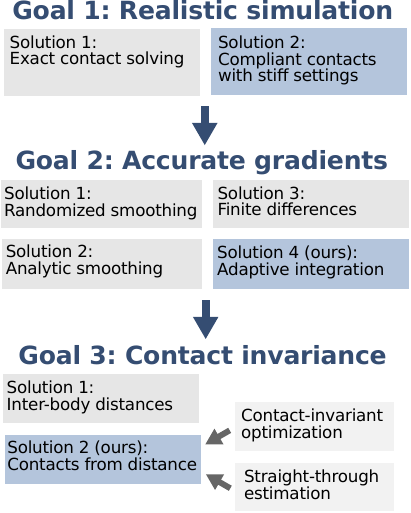}
    \caption{Overview on hierarchical goals that need to be accomplished to use simulator gradients for robot controller synthesis.}
    \label{fig:related-work-overview}
\end{wrapfigure}

Differentiable simulators are an active area of research spanning multiple fields of physics, including elastic object and fluids modeling \cite{hu2019chainqueen, warp2022, hu2020difftaichi, liu2024softmac} and modeling rigid body collisions \cite{todorov2012mujoco, hu2020difftaichi, howell2022dojo, mjwarp2025} (see \cite{newbury2024reviewDiffSimulators} for a recent overview). As outlined in  \cref{fig:related-work-overview}, we focus on robotics simulators involving hard contact collisions, with the hierarchical objectives of \myemph{(i)} accurately simulating dynamics, \myemph{(ii)} computing correct simulator gradients, and \myemph{(iii)} obtaining informative gradients between non-colliding objects.

\paragraph{Simulating contact dynamics.} Commonly deployed robot simulators can be categorized according to their contact model into impulse-based \cite{catto2018box2d, brax2021github, hu2020difftaichi}, complementary-based \cite{howell2022dojo, werling2021nimble}, and \convex \cite{todorov2012mujoco} approaches. 
Due to their simplicity and speed, impulse-based simulators \citep{catto2009modeling}, such as those using DiffTaichi \citep{hu2019taichi, hu2020difftaichi}, are widely used in game development but remain currently uncommon in robot controller synthesis.
Complementary-based contact models compute constraint forces as a solution to a constrained optimization problem either in the form of a nonlinear-complementary problem (NCP) \citep{howell2022dojo} or linear-complementary problem (LCP) \citep{tedrake2019drake, coumans2021pybullet, heiden2021neuralsim, yang2024jade}. Exactly solving NCP is an NP-hard problem \cite{kaufman2008staggered} that is considerably difficult to solve even approximately. The NCP formulation in Dojo is physically more accurate than LCP formulations \cite{howell2022dojo}, but lacks support for parallel computation limiting its utility for controller synthesis. That said, it is currently not clear which complementary formulation is best suited for robotics.

To improve computational efficiency, \mujoco \citep{todorov2012mujoco, todorov2014mujococonvex} reformulates the complementary constraint problem as a convex optimization problem. This reformulation builds on prior work on complementarity-free approaches \citep{todorov2011mujoco, drumwright2011modeling}. 
Briefly, contact forces are computed by solving a global optimization problem, in which the ability of a constraint to generate force is modulated by the geometric penetration depth at the contact interface. As an alternative, Drake \cite{tedrake2019drake} also adopts a soft patch contact model \cite{elandt2019pressure, pang2023quasidynamic}. In turn, these approaches to contact resolution soften the dynamics.

\paragraph{Contact-invariant optimization.}
To guide an optimizer toward using body collisions for task facilitation, recent works in controller synthesis and path planning incorporate inter-body distances into the optimization objective. In RL, \cite{zhang2023simulation} added inter-point distances to rewards using a softmax function. Alternatively, the algorithms in \mujoco MPC \citep{howell2022mjpc} rely on inter-point distances combined with trajectory samples establishing contact to inform the optimization. However, depending on the task at hand, setting up distance-based loss terms to inform an optimizer about the necessity for specific object collisions can quickly become cumbersome. As an alternative approach, \cite{mordatch2012contact} proposed the framework of \emph{``contact-invariant optimization''} (CIO) in which a simulation is adjusted to also apply contact forces between non-colliding bodies. While the simulation can apply non-physical contact forces, the optimization is encouraged through the addition of several loss terms to minimize the usage of these forces. In \cite{mordatch2012discovery}, CIO was extended to in-hand manipulation of objects. Our proposed extension of \emph{contacts from distance} (CFD) is inspired by CIO. Yet, while CIO applies artificial contact forces in the forward simulation, CFD resorts to the \fbtrick to only exert contact forces in a simulation that is solely used for gradient computation. In turn, CFD does not require the addition of regularization terms to a loss when used for planning. Recently, \cite{beker2025smooth} proposed soft signed distance fields as an alternative geometry representation, while also adjusting collision detection and contact force computation to be inherently soft. This approach which routes at the geometry level of robot simulation can be seen as an alternative avenue towards CIO.

 CFD's mechanism to create ``forces from a distance'' (FFD) is also conceptually similar to \cite{pang2023quasidynamic}. Drawing inspiration from works on creating FFD, this work smoothens the contact model of a quasi-static penalty-based simulator using log-barrier functions and analyzes FFD in the context of quasi-static dynamics and implicit differentiation. In comparison, CFD uses MuJoCo's smooth contact model based on polynomial splines to obtain forces from a distance. We note that both quasi-static dynamics and implicit differentiation are useful extensions to DiffMJX and CFD.

\paragraph{Straight-through estimation.} The use of the \fbtrick in CFD is inspired by a plethora of works in robotics and beyond \citep{bengio2013estimating,sahoo2023backpropagation}. 
The code underlying the \fbtrick is reported in JAX \citep{deepmind2020jax} documentation under the collective heading of \emph{straight-through estimation}\urldef\urlst\url{https://docs.jax.dev/en/latest/advanced-autodiff.html#straight-through-estimator-using-stop-gradient}.
Straight-through estimation has been used in \cite{gumbsch2021sparsely}, where the backward pass of a Heaviside function is replaced with an identity function to obtain sparse RNNs for model-based RL. \citet{horuz2025resurrection} demonstrate that using a ReLu activation with a custom backward function rivals smooth ReLu surrogates.
Recently, \cite{song2024diffsimbackward} replaced simulator dynamics obtained from IsaacGym \citep{liang2018gpuacceleratedroboticsimulationdistributed} with single-rigid body dynamics for the gradient computation in the backward pass. This approach is closely related to our work, but unlike CFD, \cite{liang2018gpuacceleratedroboticsimulationdistributed} does not utilize the forward pass simulation in the backward pass, limiting its utility for complex control synthesis tasks and system identification.

\vspace{2cm}

\begin{figure}[h]
    \centering
    \includegraphics[width=0.8\linewidth, trim={0 0 0 4mm}, clip]{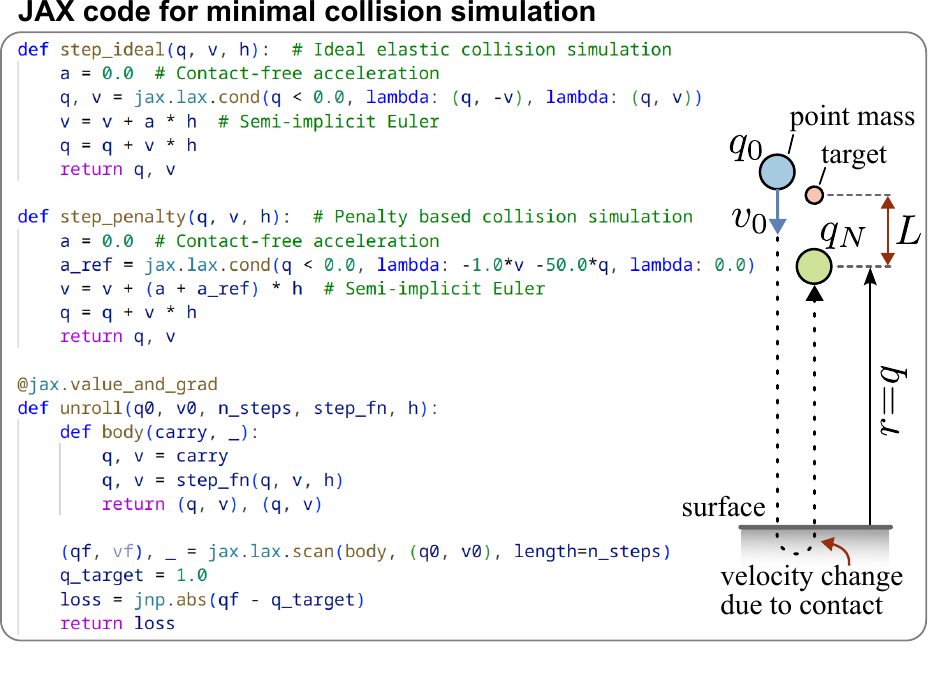}
    \caption{\myemph{Jax code for minimal collision simulation.} The code simulates a minimal version of a \convex or ideal-elastic simulator and is used in \cref{fig:minimum-toy-problem} to illustrate the TOI discretization errors as shown in \cref{fig:minimum-toy-problem}.}
    \label{fig:code-toi-toyexample}
\end{figure}

\begin{figure}
    \centering
    \includegraphics[width=\linewidth, trim={120mm 0 230mm 0},clip]{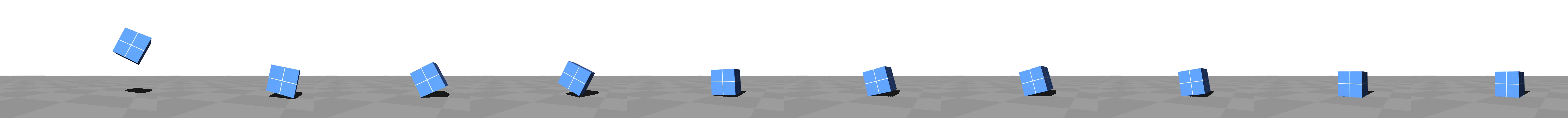}
    \vspace{-1mm}
    
    \includegraphics[width=\linewidth]{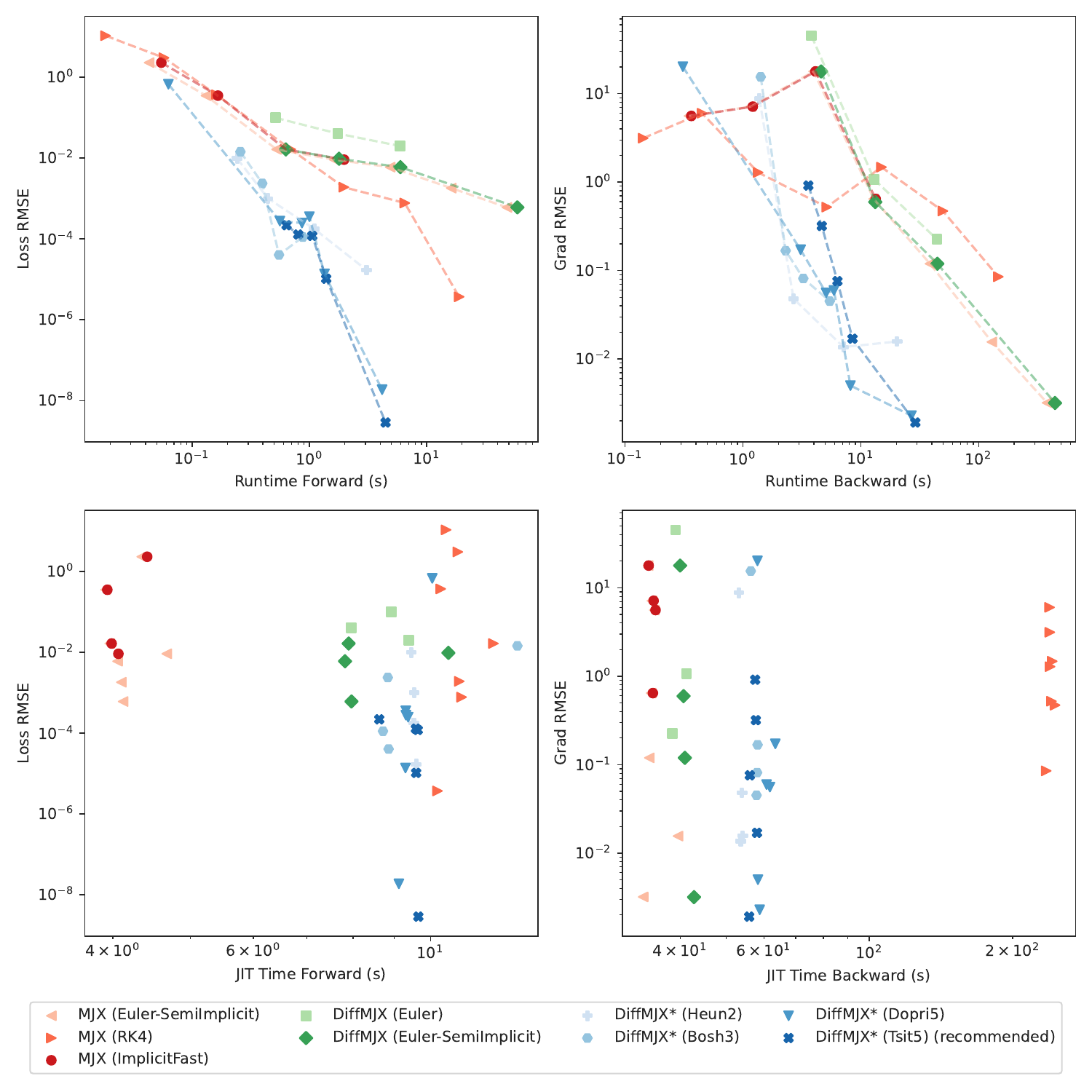}
    \caption{\myemph{Error of loss and gradient vs runtime and compilation time for different integrators.} The loss and gradient are computed for the cube toss at an initial velocity of $v_x=-2.0$ with contact settings \lstinline|solref=[0.005 1.0]| and \lstinline|solimp=[0.0 0.95 0.001 0.5 2]| as depicted in \cref{fig:object-tosses-extended} and the top of this figure. 
    The ground-truth gradients are computed from finite-differences using an adaptive solver with very low tolerance ($10^{-12}$) to simulate the rollout.
    Standard fixed-stepsize MJX integrators have red colors, fixed-stepsize integrators in Diffrax have green color, and adaptive-stepsize integrators in Diffrax have blue color (also marked by $*$).
    For adaptive stepsize control we use the Diffrax PID controller with $P=0.2, I=0.4, D=0.0$, as recommended for stiff ODEs.
    We observe that the pareto-front of adaptive solvers is shifted, allowing lower loss and gradient errors with less runtime.
    The best-performing solver is the Tsit5 solver \citep{tsitouras2011runge}, which is a 5th order explicit Runge--Kutta method with an embedded 4th order method for adaptive step sizing.
    This is also the default solver used in other experiments.}
    \label{fig:object-tosses-scatter}
\end{figure}

\newpage
\section{Tuning DiffMJX}
\label{sec:appendix:tuning-diffmjx}
\begin{figure}[t]
    \centering
    \includegraphics[width=0.7\linewidth]{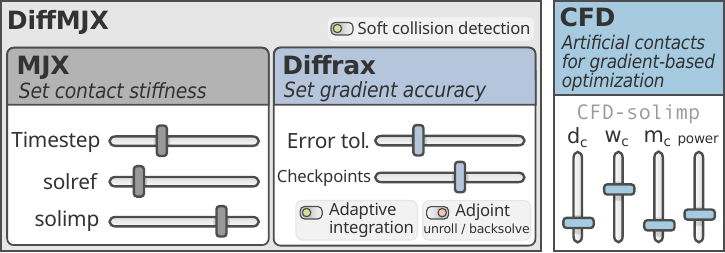}
    \caption{\myemph{DiffMJX and CFD add tuning knobs to MJX's gradient computation.} DiffMJX adds support for adaptive integration atop MJX. The integrator's error tolerances trades computational speed for improved gradient accuracy. The number of checkpointing steps trades computation speed with GPU memory consumption. CFD adds parameters that set the distance and magnitude at which artificial contact forces between non-colliding objects are applied in the gradient computation.}
    \label{fig:tuning_knos_diffmjx}
\end{figure}

Through Diffrax, DiffMJX provides a whole suite of additional tuning knobs that can be used to improve gradient computation as shown in \cref{fig:tuning_knos_diffmjx}.

\paragraph{Error tolerances set gradient accuracy.} 
In this work, we showed that the errors of penalty-based simulator gradients depends on
\begin{enumerate}[noitemsep]
\vspace{-2mm}
\item the \myemph{simulation stiffness}, aka the hardness of contacts and joint limits, 
\item the \myemph{contact velocities} in normal direction, aka how fast objects collide into each other,
\item and the \myemph{integration stepsize}.
\vspace{-2mm}
\end{enumerate}

While simulation stiffness and contact velocities are system dependent, the integration stepsize is a tuning parameter. To obtain accurate gradients, one can use a constant stepsize integrator with a sufficiently small stepsize which inevitably results in slow integration. 

To speed up computation, an adaptive integrator solely reduces the stepsize if necessary as a function of the integrator's error tolerances.
In turn, as the integrator's error tolerances directly determines the gradient errors, it forms the most important hyperparameter.
\begin{quote}
\vspace{-2mm}
\emph{To speed up computation, use the largest error tolerance that still produces sufficiently accurate gradients.}
\vspace{-2mm}
\end{quote}

\Cref{fig:object-tosses-scatter} shows the error vs runtime trade-off for different integrators in MJX and DiffMJX.
We observe that in the case of the cube toss, \textbf{the adaptive integrators reduce the runtime required to achieve gradients with low error}.

\paragraph{Number of checkpoint trades GPU memory for compute time.}
Thanks to Diffrax's advanced techniques for checkpointing, DiffMJX memory consumption is significantly reduced compared to MJX. As noted in Diffrax's documentation, the memory used approximately equals the number of checkpoints multiplied by the size of the ODE's state. 
\begin{quote}
\vspace{-2mm}
\emph{Increasing the number of checkpoints speeds up computation. Therefore, use as many checkpoints as possible given the available GPU memory.}
\vspace{-2mm}
\end{quote}
By reducing the number of checkpoints, we were able to run DiffMJX simulations that MJX could not, because MJX exceeded the available GPU memory.

\paragraph{Maximum number of ODE steps affects JIT compilation time.} 
As shown in \cref{fig:object-tosses-scatter}, there is a notable increase in JIT compilation time when using Diffrax compared to \mujoco's semi-implicit Euler integrator. This is due to the program being compiled for the maximum number of ODE steps that is set by the user. Therefore, this parameter should be chosen as small as possible. 

\newpage
\section{Method Details} \label{sec:additional-remarks}

\subsection{Removing discontinuities from collision detection and muscle actuators} \label{sec:appendix:collision-detection}

The collision checking in MJX is implemented for pairs of several different geometries.
Many of the collision check implementations rely on several case distinctions, e.g. the collision of a cylinder and a plane is separated into the case of the cylinder being parallel to the plane vs not.
These hard case distinctions can lead to errors in the gradients, as they introduce discontinuities in the dynamics and its gradient.
We adopt the typical approach for smoothing such non-differentiabilities by replacing any hard case distinctions by smoothly interpolating between the cases with weights computed from a sigmoid.
Specifically, we end up modifying the collision detection between plane-cylinder, sphere-capsule, capsule-capsule, plane-capsule and plane-box pairs.
More complicated collisions like mesh-mesh collisions are currently not modified as they are not relevant in our experiments, but we plan on supporting these in a future version.

As an example, consider our soft version of a function for plane-box collision detection. More details on other smoothed functions will be available in the released codebase.
\lstset{
    basicstyle=\tiny\ttfamily\color{blue},  }
\begin{lstlisting}
def _plane_box(plane: GeomInfo, convex: ConvexInfo) -> Collision:
  """Calculates contacts between a plane and a convex object."""
  vert = convex.vert

  # get points in the convex frame
  plane_pos = convex.mat.T @ (plane.pos - convex.pos)
  n = convex.mat.T @ plane.mat[:, 2]
  support = (plane_pos - vert) @ n
  # search for manifold points within a epsilon skin depth
  threshold = jp.maximum(0, support.max() - 1e-3)
  soft_poly_mask = math.sigmoid(support, low=threshold - 1e-3, high=threshold)
  support_masked = math.soft_where(soft_poly_mask, support, 0)

  dist_neg, idx = jax.lax.top_k(support_masked, 4)
  dist = -dist_neg
  pos = vert[idx]

  # convert to world frame
  pos = convex.pos + pos @ convex.mat.T
  n = plane.mat[:, 2]

  frame = jp.stack([math.make_frame(n)] * 4, axis=0)
  pos = pos - 0.5 * dist[:, None] * n
  return dist, pos, frame

def soft_where(condition, x, y):
  return condition * x + (1 - condition) * y
\end{lstlisting}

We also encountered differentiation errors in MJX environments with muscle–tendon units.
The source of these issues is that the functions for computing wrapping of muscles around objects rely on nonlinearities such as arcsin, which has an undefined gradient at some points.
A practical remedy is the “double-where” trick, which prevents gradient flow at those singularities.

Concurrently with our work, \cite{la2025motion} employed MJPC \cite{howell2022mjpc}  -- a library for CPU-based MPC in MuJoCo via finite differences -- to compute ideal controls of a musculoskeletal dog model. This seminal work also reports that modifying the muscle dynamics to be differentiable was a key contribution for improving MPC performance.

\subsection{ODE differentiation in Diffrax} \label{sec:appendix:diffrax}

\paragraph{Diffrax: Numerical integration with JAX} \label{sec:appendix-integration-with-diffrax}
The ODE solved by the Mujoco simulator can be written as
\begin{align}
    x(0) &= x_0
    &
    \dot{x}(t) &= F_\theta(t,x(t)),
\end{align}
where $\theta$ incorporates all model parameters and $x$ is the full state of the system.

In this work, we resort to the Diffrax library \citep{kidger2021on,kidger2021equinox} for numerical integration in Jax to solve the above ODE. 
This powerful library provides efficient implementations of many fixed and adaptive timestep solvers \citep{tsitouras2011runge,dormand1980family}.
It also allows easily switching between different modes for backpropagation: Discretize-then-optimize (also referred to as unrolling) and optimize-then-discretize (also referred to as backsolving) \citep{chen2018neuralode,kidger2021on}.
Overall, the choice of solver, stepsize controller and differentiation technique is typically application-dependent.
Hence, we implement the general Diffrax integrator as an easy-to-use alternative for the existing fixed-stepsize integrators in MJX, offering the full flexibility of the Diffrax library to the user.
Notably, \emph{we adjust the Diffrax solvers to accommodate for exact integration of quaternions and stateful actuators}, similar to the Runge-Kutta implementation of MJX, which reduces the number of required integration steps.

\paragraph{Discretize-then-optimize.}
The first notable differentiation mode is discretize-then-optimize.
It is the result of discretizing the forward ODE with a numerical integrator and then computing the gradients of the discretized forward ODE by unrolling the computation graph.
This approach aligns with the paradigm of differentiable programming used in autodifferentiation libraries such as JAX, hence this is also the approach taken in standard MJX.
The problem here is that the memory requirements scale linearly with the number of solver steps. 
For many applications this is a serious bottleneck in standard MJX, and the problem is exacerbated when relying on adaptive integration, as it typically involves storing even more substeps.
Fortunately, Diffrax implements an optimal gradient checkpointing scheme that allows reducing memory requirements from $O(n)$ to $O(1)$ in exchange for increasing runtime from $O(n)$ to $O(n\log n)$, where $n$ is the maximum number of solver steps.

Another potential issue is that discretize-then-optimize inherently computes gradients of the discretized dynamics, rather than the true continuous dynamics.
This can lead to interesting failure cases in which even in the limit of the stepsize going to zero, the computed gradient is different from the true gradient of the continuous dynamics.
This is the result of the basic fact that the convergence of a parameterized function (the discretized ODE) to a limit (the continuous ODE) does not imply the convergence of its gradient.
Note that this is not a mere mathematical corner case, instead it is the underlying reason for time-of-impact oscillations as described in \cref{fig:minimum-toy-problem} and \cite{hu2020difftaichi, schwarke2024learning}.

\paragraph{Optimize-then-discretize.}
Fixing this requires (approximately) computing the gradients of the true continuous forward ODE.
This is achieved by analytically computing the gradient as the solution to the continuous adjoint equations.
Assuming a loss $L=L(x(T))$ only on the final state w.l.o.g., the continuous adjoint equations \citep[Theorem 5.2]{kidger2021on} can be written as:
\begin{align}\label{eq:adjoint-equations}
    a_x(T) &= \frac{dL}{dx(T)}
    &
    \dot{a}_x(t) &= -a_x(t)^\top\frac{\partial F_\theta}{\partial x}(t,x(t))
    \\
    a_\theta(T) &= 0
    &
    \dot{a}_\theta(t) &= -a_x(t)^\top\frac{\partial F_\theta}{d\theta}(t,x(t)).
\end{align}
Solving these using any numerical solver on the backward pass, as offered readily by the Diffrax library, yields the desired gradients $\frac{dL}{dx(t)}=a_x(t)$ and $\frac{dL}{d\theta}=a_\theta(0)$.

This approach is called ``optimize-then-discretize'' or BacksolveAdjoint in Diffrax.
Note also that the memory requirements can again be reduced to constant in the number of solver steps, by solving the forward ODE ``backwards in time'' together with the adjoint ODE.
Optimize-then-discretize, in contrast to unrolling, allows to directly specify tolerances on the error in the gradients.

One of the intuitive potential benefits of optimize-then-discretize is that the adaptive stepsize controller for solving the adjoint ODE now has the ability to adapt the stepsize when computing the adjoint derivatives, rather than being constrained to unrolling the steps that were taken during the forward solve.
However, in practice it is typically the case that the forward ODE and the adjoint ODE are stiff ``in the same regions'', i.e. the state derivative changes quickly when the state itself changes quickly.
Hence, when unrolling the forward solve, the steps are already small in the regions where the adjoint ODE is also stiff.
Therefore it typically suffices to use the less complicated discretize-then-optimize, which has a less complex computational graph and therefore comes with lower compilation times.
The above explanation of course only holds when the forward ODE and the adjoint ODE ``match''; when modifying the adjoint ODE (as will be discussed in the following section) it may be beneficial to use optimize-then-discretize.

\newpage
\subsection{Theoretical results on adaptive ODE integration}
There exists a rich history of results on analyzing discrete adjoint sensitivities, supporting the claims on gradient accuracy improvements from non-adaptive solvers.
For instance, \cite{sandu2006rungekutta,walther2007automatic} establish that for one-step Runge–Kutta methods of order-p, the reverse-mode sensitivities obtained by differentiating the discrete solve are themselves order-p consistent and stable. This directly links reductions in forward global error to reductions in gradient error at the same order.
\cite{calvo2008global} show that (under standard Lipschitz/stability conditions) the global state error scales like the (relative/absolute) tolerance enforced by the controller; the adjoint/sensitivity error inherits this scaling with constants depending on the assumed Lipschitz constant. As our setting is an application of adaptive integration to the ODE given by the simulator, these results apply out of the box and establish a theoretical basis for the observed improvements in gradient accuracy from the use of adaptive solvers. We hope that these changes will strengthen the presentation.

The advantages and disadvantages of discretize-then-optimize (DTO) vs optimize-then-discretize (OTD) have also been researched for many years. This includes \cite{hinze2008optimization} comparing DTO and OTD, giving conditions for equivalence, and providing error bounds in the discretized optimality system.
\cite{hager2001rungekutta} proves accuracy of discrete optimality systems under Runge-Kutta time discretization and establishes conditions ensuring correct gradients.
Moreover, \cite{bonnans2006computation} provide order conditions for (partitioned) Runge–Kutta schemes in optimal control and implications for obtaining matching accuracy in the adjoint/gradient after discretization.

\begin{figure}[t]
\vspace{-12mm}
\begin{minipage}[t]{0.33\textwidth}
    \centering
    \includegraphics[width=\linewidth]{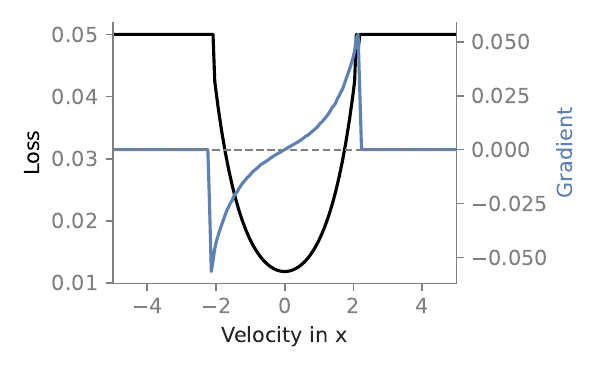}
\end{minipage}
\begin{minipage}[t]{0.33\textwidth}
    \centering
    \includegraphics[width=\linewidth]{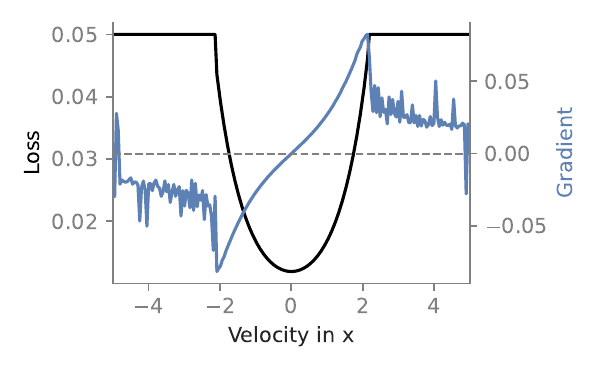}
\end{minipage}
\begin{minipage}[t]{0.33\textwidth}
    \centering
    \includegraphics[width=\linewidth]{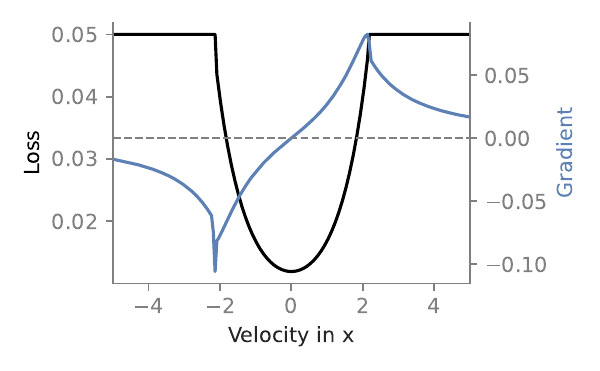}
\end{minipage}
\vspace{-7mm}
\caption{
\myemph{Optimize-then-discretize vs discretize-then-optimize in the presence of CFD.}
Results are for a reduced billiard example as in \cref{fig:billiard_loss} without table-ball contacts and gravity.
\myemph{Left:} DiffMJX adaptive integration results in well-behaved gradients, but gradient is zero if objects do not collide.
\myemph{Middle:} Using CFD with \emph{discretize-then-optimize} results in small gradient oscillations. The adaptive integrator selects large stepsizes as no contacts are happening (forward ODE is non-stiff), whereas in the gradient computation the CFD are added (adjoint ODE is stiff). 
In turn, unrolling the large stepsizes cause discretization errors.
Note that this phenomenon persists no matter how low the tolerance of the forward ODE solver is set, as the stepsize controller is unaware of the CFD mechanism that is only used in the gradient computation.
\myemph{Right:} Using \emph{optimize-then-discretize} allows the solver for the adjoint ODE to select small stepsizes when the adjoint ODE is stiff due to CFD. This fixes the gradient oscillations.
}
\label{fig:billiard_cfd_unroll_vs_backsolve}
\vspace{-2mm}
\end{figure}

\newpage
\subsection{Contacts from distance with the \fbtrick} \label{sec:appendix:fbtrick}
In discretize-then-optimize, automatic differentiation automaticaly uses the vector-jacobian-products derived from $\tilde F_\theta$ instead of $F_\theta$.
In optimize-then-discretize, with the straight-through-trick applied to $F$, we replace the vector-jacobian-products in the adjoint equations \cref{eq:adjoint-equations} as follows
\begin{align}
    \tilde{a}_x(T) &= \frac{dL}{dx(T)}
    &
    \dot{\tilde{a}}_x(t) &= -\tilde{a}_x(t)^\top\frac{\partial \tilde{F}_\theta}{\partial x}(t,x(t))
    \\
    \tilde{a}_\theta(T) &= 0
    &
    \dot{\tilde{a}}_\theta(t) &= -\tilde{a}_x(t)^\top\frac{\partial \tilde{F}_\theta}{d\theta}(t,x(t)).
\end{align}
Solving these altered adjoint equations gives replacements for the respective gradients, which in our case incorporate gradient information on contacts from distance.
Again, this approach does not require any additional implementation effort, as the straight-through-trick allows to just use the existing Diffrax implementation of optimize-then-discretize.

\paragraph{Optimize-then-discretize vs Discretize-then-optimize in the presence of CFD.}
As discussed in the previous section, adding contacts from distance may change the preferred choice of gradient computation, as the forward and adjoint ODE may have different stiffness values in different regions.

To illustrate this, we again consider a version of the billiard-toy example from \cref{fig:billiard_loss}.
This time, we turn gravity off and disable the table-ball contacts to isolate the ball-ball contact, which means that the ODE is linear before and after the collision.
We again compute the loss and gradient over a range of initial parameters, the results are reported in \cref{fig:billiard_cfd_unroll_vs_backsolve}.
As before, we observe that with the DiffMJX adaptive integrator, we get well-behaved gradients, but the gradient becomes zero when the two balls do not collide.

Now, we activate the CFD mechanism and repeat the experiment. In this case, the gradient show oscillations due to the different stiffness settings of the forward and adjoint ODE:
The solver for the forward ODE takes very large steps when no ball-ball collision happens, but the adjoint ODE incorporates the collision signal and hence is stiff. 
Simply unrolling the forward integration therefore makes a discretization error, resulting in the oscillations.
Note, that we did not observe this phenomenon in \cref{fig:billiard_loss}, as this experiment includes ball-table collisions which cause the adaptive solver of the forward ODE to take small steps even when no ball-ball collision is happening.
The solution is to use optimize-then-discretize. 
Here, the adaptive solver for the adjoint ODE can select stepsizes according to the stiffness of the adjoint ODE.
This is also confirmed by the results in \cref{fig:billiard_cfd_unroll_vs_backsolve} (right).

This experiments highlights a specific corner case, and we did not observe issues in our other experiments when using discretize-then-optimize with CFD, as we typically do not have such extreme variations of the stiffness in the forward ODE.
However, we believe it is beneficial to be aware of this potential caveat and how one can resolve it with the tools available in DiffMJX.

\section{DiffMJX \& CFD: The sharp bits}
\label{sec:app-sharp-bits}

\paragraph{CFD adds additional contacts to the solver which can slow down computation.} The documentation of \mujoco XLA\footnote{\url{https://mujoco.readthedocs.io/en/stable/mjx.html\#mjx-the-sharp-bits}} advises to use meshes with 200 vertices or fewer due to computational limitations on the current implementation of collision detection. If MJX is used with CFD, then every contact is added to the contact solver for which $r<w_c$. In turn, for dense meshes and large $w_c$, computation time notably increases. The parameter $w_c$ can be made arbitrarily small such that, in its limit for $w_c=0$, we retain vanilla MJX with our improvements on the differentiability of the collision detector. 
That said, for in-hand manipulation tasks and robot locomotion tasks, the number of objects colliding at one instance is small enough to render CFD a useful tool for easing optimization with \convex simulators.

\paragraph{Adaptive integration slows down training if tolerances are set too small.} 
For the very large initial cube side length initial value of $140$ mm, DiffMJX saw a significant drop in computation speed forcing us to abort these runs. This is not surprising as penetrations amounting to 40\% of the total cube's width cause huge contact forces that significantly stiffen the ODE. 
In Appendix \ref{sec:appendix-sysid}, we reduce the integration error tolerance which allowed optimization to converge even for large initial parameters.
Alternatively, one could speed up training for these too large geometry parameter initializations by first using soft impedance settings that during training becomes annealed to become increasingly stiff. 
The same principle applies to  \emph{scenes with multiple objects}: overly tight integration error tolerances can substantially slow down the simulation. 
As integration error scales with relative contact velocity, scenarios involving large state changes along contact normals trigger smaller step sizes. Notably, while the ball in the billiard shot example (\cref{fig:billiard_loss}) was rolling over the table, the integration stepsize did only marginally decrease. Whereas during the collision between balls, the integrator adjusted the stepsize notably.

\paragraph{Finite differences can be a viable alternative for computing low-dimensional gradients.} In general, gradient computation with MJX via automatic differentiation is notably slower than running only the forward simulation.
This makes zeroth-order methods such as predictive sampling favorable whenever the task can be solved with them. This is usually the case for low-dimensional systems or computing gradients of only a few parameters, e.g. compting gradients via finite differences for less then ten parameters seems to work well in practice. 
As sampling suffers from the curse of dimensionality, automatic differentiation for computing gradients scales to extremely high-dimensional control problems which enabled us to do gradient-based MPC on the musculoskeletal systems.

\paragraph{CFD should be combined with sampling to solve nonconvex optimizations.} For MPC, our CFD mechanism tends to be effective when the object needs to be pushed in a specific direction by the hand, but it performs poorly in grasping tasks. This limitation arises because, at a distance, the cumulative force arising from the hand's CFDs conglomerate into a force that pushes objects away.
However, the gradient does not encode the possibility of grasping, as this signal only emerges when the object is inside the hand. Fundamentally, this issue reflects the non-convex nature of the optimization landscape.
However, we do not believe that this is a problem that should be solved using a CFD mechanism; rather, in such tasks, the best approach would be to combine CFD with sampling to overcome the non-convexity while maintaining the favorable scaling of gradients with dimensionality.

\begin{figure}[tb]
\vspace{-8mm}
\centering
\includegraphics[width=0.9\linewidth]{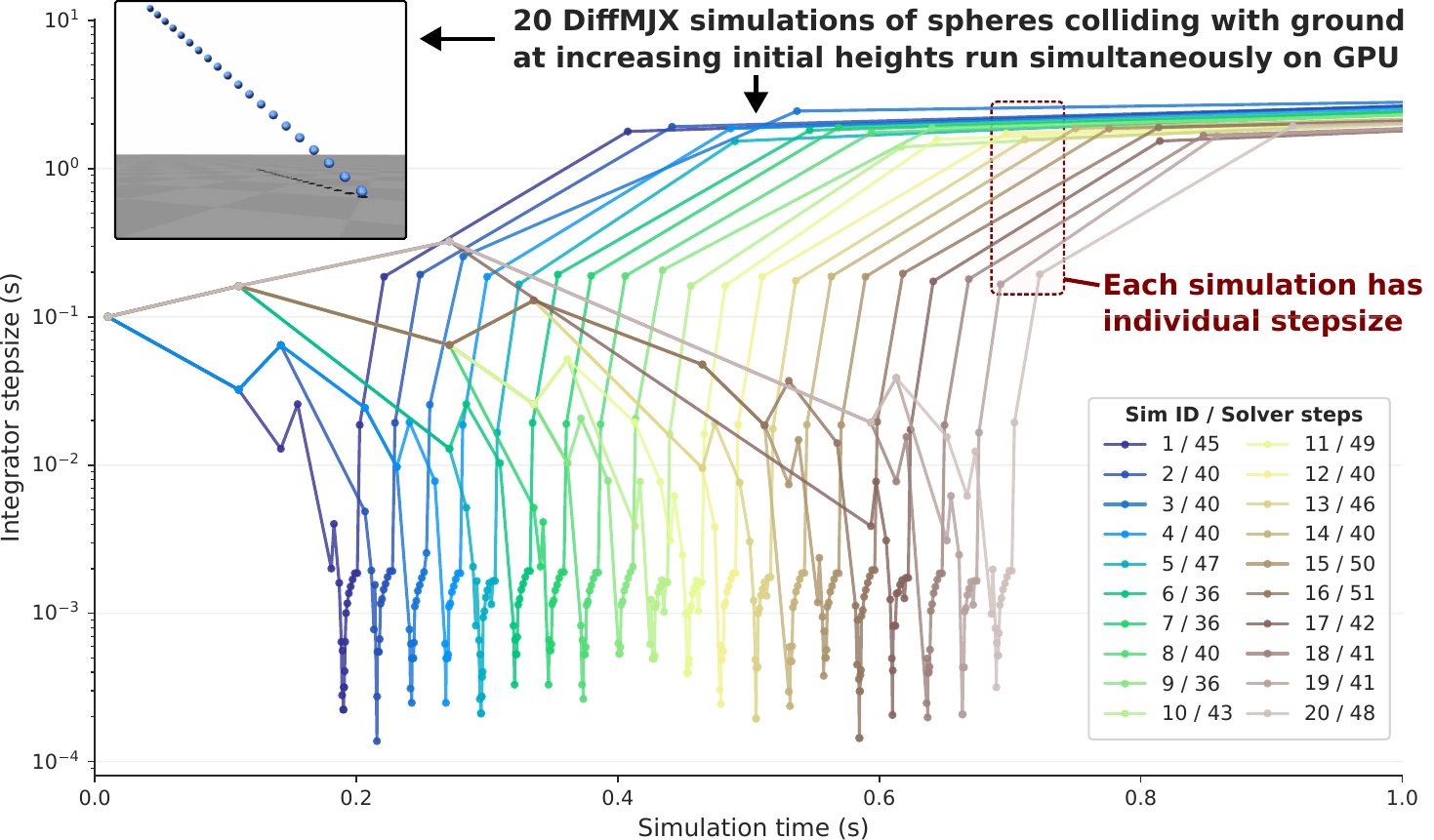}
\caption{We vmap the DiffMJX simulation of a ball drop across 20 different environments. In each environment, a single ball is dropped (with an initial downwards velocity and no gravity) onto a plane, from which it bounces back up. The 20 environments start with the ball at different heights, such that the contact happens at a different simulation time for each environment. We then plot the selected adaptive stepsize of the solver over the simulation time for each environment, and observe that the stepsize is indeed independent for each run, with a reduction in stepsize only occurring at the single contact in a specific environment. Therefore, when vmapping across environments, contacts in one simulation do not slow down other simulations.}
\label{fig:vmap_ball_drop}
\end{figure}

\paragraph{Parallelizing DiffMJX simulations.}
Locomotion policies trained with reinforcement learning typically rely on domain randomization with many environments executed in parallel. DiffMJX supports such batched execution via JAX’s vmap, analogous to MJX.
However, if a single policy induces large contact forces, then this may trigger very small integration steps, which results in the other simulations to wait for a slow simulation to finish. This slowdown is mitigated by increasing the integration error tolerances or by early-terminating the respective simulation once a predefined solver-step budget is exceeded.
However, it is important to note that such a slowdown is only observed when many contacts are happening in a single environment. 
On the other hand, when few contacts are happening in each parallelized environment, there is no additional slowdown due to vmapping. This is because the selected adaptive stepsize in Diffrax is not global across the vmapped environments, it is fully vmapped and can take different values across environments. We experimentally verify this in \cref{fig:vmap_ball_drop}, see the caption for details.

\section{Evaluation details and further experiments} \label{sec:appendix-experiment-details}

\subsection{Gradient analysis of MJX} \label{sec:appendix-gradient-analysis}
\begin{figure}[t]
\vspace{-4mm}
    \centering
    \includegraphics[width=\linewidth]{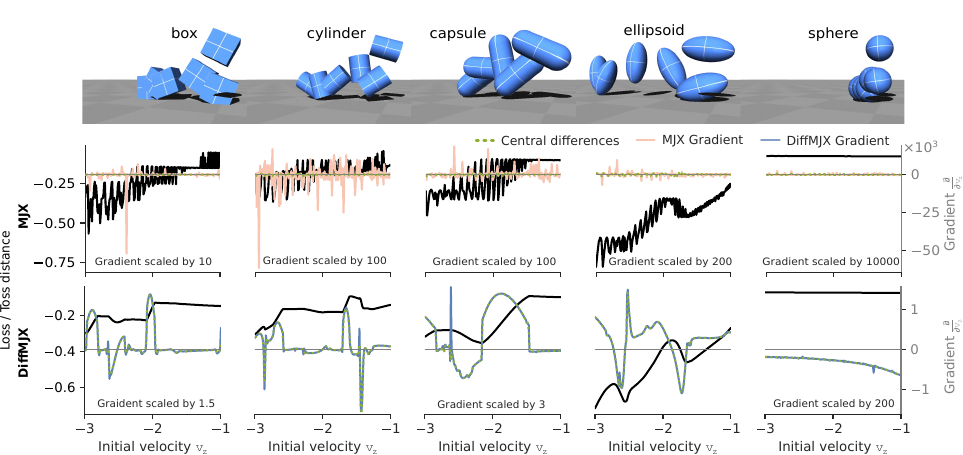}
    \caption{Simulation of geometric primitives thrown onto a surface using MJX or \method. Contacts at default stepsize $0.002\,\mathrm{s}$, \lstinline|solref=[0.005 1.0],| and \lstinline|solimp=[0.0 0.95 0.001 0.5 2]| cause MJX's gradients of the toss distance to deviate significantly from central difference gradients, while \method maintains close agreement.}
    \label{fig:object-tosses-extended}
\end{figure}

In \cref{fig:minimum-toy-problem}, the time discretization errors arising in \convex simulators are first illustrated on a minimal toy example. The code to generate this toy example is shown in \cref{fig:code-toi-toyexample}. After illustrating on this toy example that time discretization errors can be mitigated by reducing the stepsize, it is shown in \cref{fig:object-tosses} that these errors also occur in \mujoco. \cref{fig:object-tosses-extended} is an extended version of \cref{fig:object-tosses} containing all of \mujoco's basic shape primitives. 

\subsection{System identification experiments} \label{sec:appendix-sysid}
\paragraph{Contactnet dataset}
The contactnet dataset\footnote{https://github.com/DAIRLab/contact-nets} consists of $550$ trajectories of an acrylic cube that has been repeatedly tossed onto a wooden table. As reported in \cite{pfrommer2021contactnets, acosta2022validating}, data has been collected at  1480 Hz and the cube's physical parameters amount to a side length of $10\,\mathrm{cm}$, mass of $0.37\,\mathrm{kg}$, inertia of $0.0081\,\mathrm{kg\,m^2}$, a friction coefficient of $0.18$, and restitution of $0.125$.

\paragraph{Training setup}
The trajectories are split into segments of length five. In the supervised train loop, given a trajectory segment's initial state, each simulator is unrolled for four steps. Subsequently, an mean square error (MSE) loss between the segment's states and predicted states is computed. Before being fed to the loss function, each state's quaternion is converted to a rotation matrix to avoid representation singularities negatively affecting training \cite{geist2023rotations, bregier2021deep}. For adaptive integration, DiffMJX is set to use ``RecursiveCheckpointing'' and a 5th order explicit Runge--Kutta method (``Tsit5'' in Diffrax) with PID error tolerances of 1e-5. We use the Adam \cite{kingma2015adam} implementation from the Optax library \cite{deepmind2020jax} for gradient-based optimization with parameters \lstinline|[b1 = 0.5, b2 = 0.9, eps = 1e-6]|. 
\mujoco's parameter are set to a timestep of 0.006767\,s (the data collection sampling time), \lstinline|iterations=4| (contact solver iterations), \lstinline|ls_iterations=10| (solver steps), \lstinline|tolerance=1e-8|, \lstinline|impratio=1.0|, \lstinline|solref=[0.02 1]|, \lstinline|solimp=[0.05 0.95 0.01 0.5 2]|, and friction \lstinline|solimp=[0.4 0.01 0.1]|. The cube's mass is set to the reported groundtruth value and its inertia is computed by \mujoco's default equal density approximation. DiffMJX uses the same \mujoco settings with the exception of \lstinline|iterations=2| (adaptive integration does not require as many solver iterations).
The CFD impedance function is set \lstinline|[0.0, 0.01, 0.01, 1.0, 4.0]| that is $d_c=0.01$, $d_0=0$ with $w_c=1\,\mathrm{m}$.

\subsubsection{Identification of additional \mujoco parameters}
In this section, we extend the experiment from \cref{sec:sysid} and use DiffMJX to  estimate \mujoco's simulation parameters alongside the cube's side length. In these experiments, the same setup is used for DiffMJX as detailed in the previous section. In comparison to the experiment in \cref{sec:sysid}, the integration error tolerance is increased to \lstinline|1e-4| to reduce computation time. Moreover, the parameters are constrained using either a softplus function or a softclip function where the softening hyper-parameter has been carefully tuned. While the models are trained on an MSE between trajectory segments, we evaluate the model on the ``trajectory error'' being the mean absolute error between the first forty trajectories in the dataset and DiffMJX's trajectory prediction. In the these experiments, DiffMJX estimates the following parameters simultaneously:

\begin{itemize}
	\item  sidelength as also estimated in \cref{sec:sysid},
	\item  mass from which \mujoco automatically computes the cube's inertia,
	\item solref parameters (time constant $t_c$, damping ratio $\phi_d$) determining the constraint stiffness, 
	\item solimp parameters determining the constraint's ability to generate force,
	\item friction parameters determining the extend of the contact friction cone.
\end{itemize}

\begin{figure}[t]
    \centering
    \includegraphics[width=0.65\linewidth]{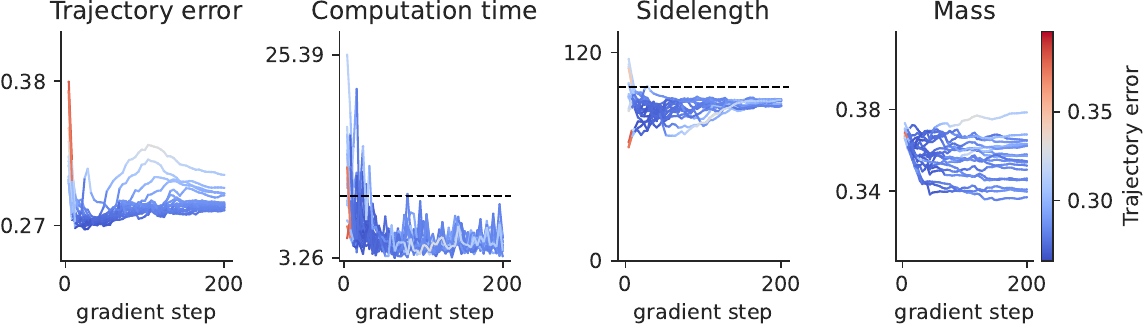}
    \vspace{3mm}
    
    \includegraphics[width=0.65\linewidth]{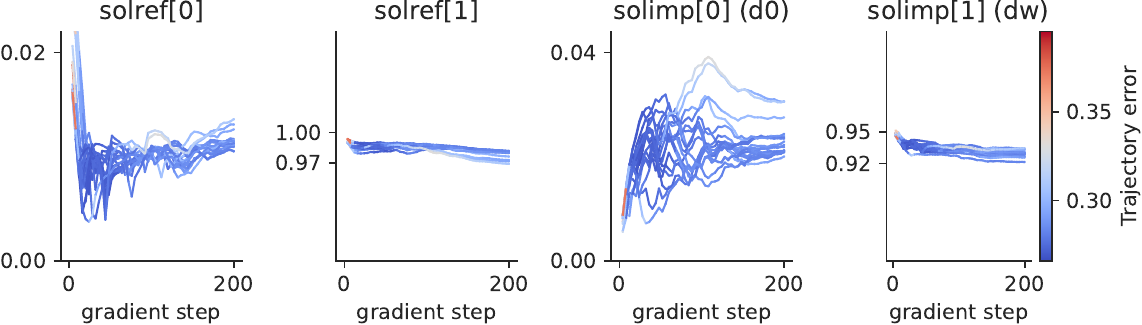}
    \vspace{3mm}
    
    \includegraphics[width=0.8\linewidth]{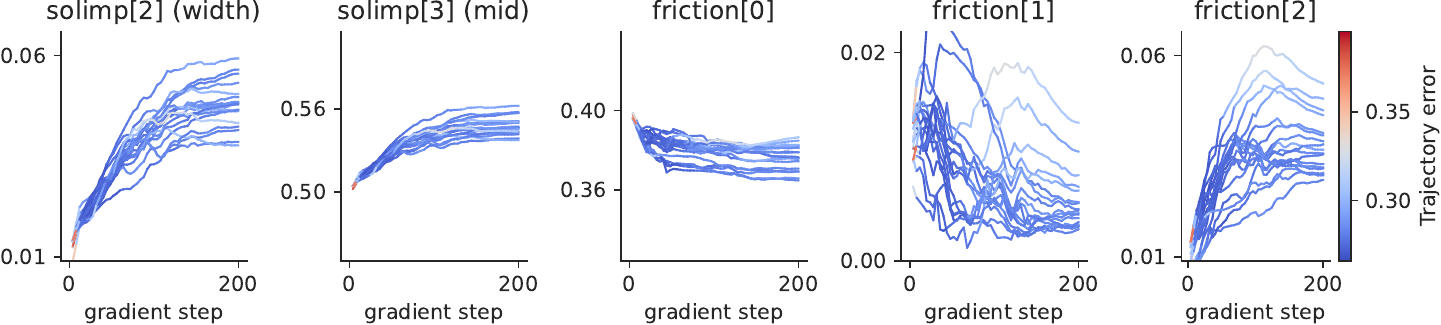}

\noindent\rule{8cm}{0.4pt}
\vspace{2mm}

    \includegraphics[width=0.65\linewidth]{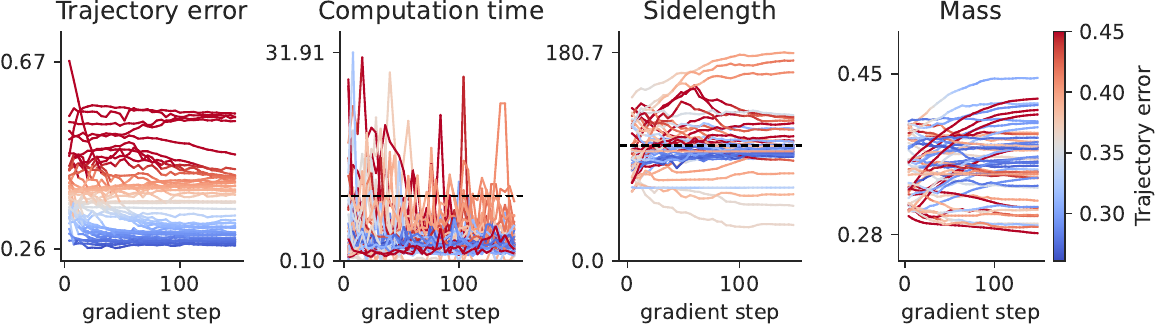}
    \vspace{3mm}
    
    \includegraphics[width=0.65\linewidth]{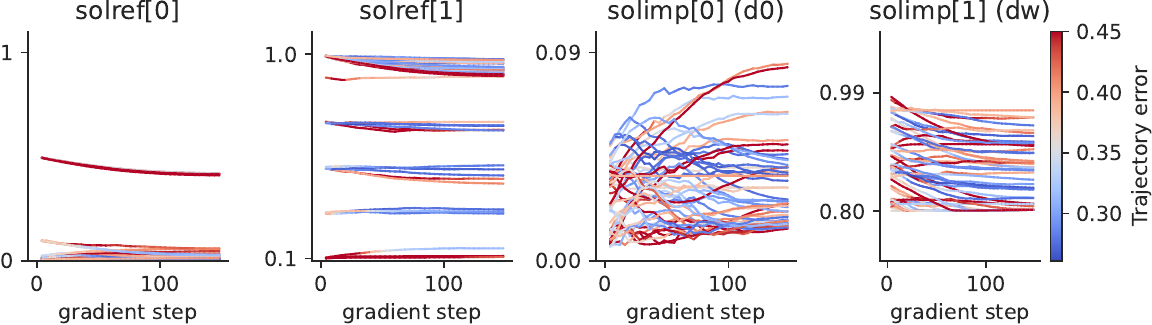}
    \vspace{3mm}
    
    \includegraphics[width=0.8\linewidth]{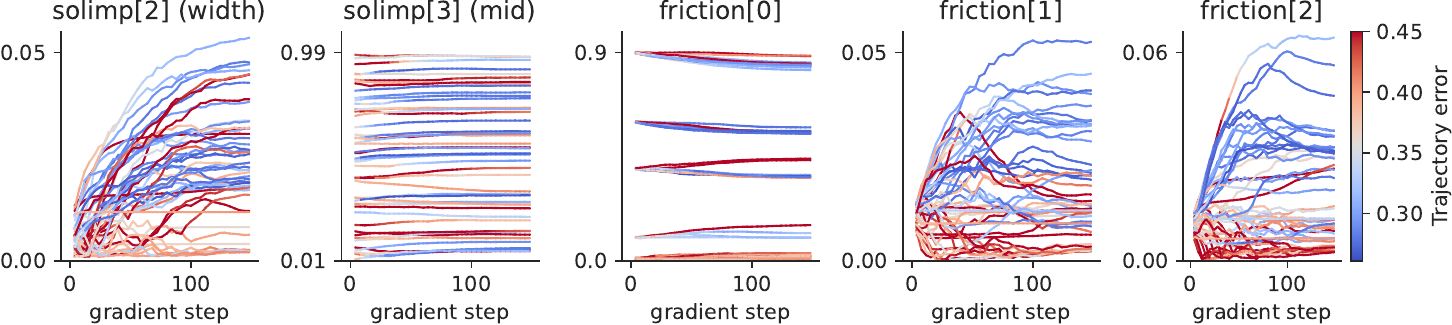}
    \caption{\myemph{Top:}  Identification of \mujoco parameters with DiffMJX on Contactnets cube toss dataset starting from pre-tuned inital conditions. \myemph{Bottom:} Identification starting from random initial parameters. The cube's geometry has a significant effect on the time of impact such that the optimization converges to a sidelength close to 100\,mm to reduce the train loss. The effect of other model parameters on the system dynamics is more inter-twined. For example, a too large mass resulting in larger penetrations during impact can be compensated by increasing the constraint stiffness via \lstinline|solref[0]| or \lstinline|solimp[0]|. }
    \label{fig:sysid_extra}
\end{figure}

In total, we conduct two additional experiments with the Contactnets dataset:

\paragraph{Starting from pre-tuned initial conditions:} The optimization is started from the same pre-tuned initial conditions as used in \cref{sec:sysid}.
The prediction horizon is set to $N=10$. The training results are shown in \cref{fig:sysid_extra} (Center). Each model requires around 10 seconds of computation time per gradient step.

\paragraph{Starting from random initial conditions:}
The optimization is started with random initial parameters. The prediction horizon is set to $N=4$ such that the majority of runs require around 5 seconds for a gradient step. The training results are shown in \cref{fig:sysid_extra} (Bottom).

In both experiments, the trajectory error remains above 0.25. As noted by \citet{parmar2021fundamental}, the dynamics of a hard cube exhibit a degree of chaotic behavior. Consequently, small variations in initial conditions -- potentially arising from sensor noise or external disturbances such as wind -- can substantially alter the cube’s trajectory after one of its corners hits the table. Nevertheless, the reduction in trajectory error from approximately 0.32 to 0.26 results in the simulated cube notably more closely matching the real-world data, as shown in \cref{fig:sysid_extra}.

\subsection{MPC experiment} \label{sec:appendix-mpc}
\paragraph{Experimental setup.}
The MyoHand and MyoArm collision parameters are set to \lstinline|solref=[0.02 1.0],| and \lstinline|solimp=[0.0 0.95 0.001 0.5 2]|, the racket collision parameters are set to \lstinline|solref=[-100000 0]| for elastic collision.
When using CFD, we set \lstinline|solimp=[0.1 0.95 0.001 0.5 2]| and use \lstinline|solimp-CFD=[0.0 0.1 1.0 1.0 4]|.
The timestep of the simulation is set to $0.0025$, the constraint solver is the Newton solver with $4$ iterations and $16$ linesearch iterations. All models use our refinements for improving differentiability of \mujoco's collision detector.

All MPC experiments are performed on a NVIDIA GeForce RTX 3060.
In the in-hand manipulation task the runtimes for the gradient-based MPC are $2.9$h ($+2.2$h JIT) without CFD and $6.4$h ($+2.1$h JIT) with CFD.
The sampling with $1024$ samples takes $35$min ($+3$min JIT).
In the bionic tennis task, the runtimes for the gradient-based MPC are $9.7$h ($+1.2$h JIT) with CFD.
The sampling based MPC runs for $1.4$h with $2048$ samples.

Note, that we did not optimize any of the methods for speed in this experiment, i.e. in baoding with gradients the task was solved after less than half the total timesteps.
The main point here is that we can solve the challenging task using a purely gradient-based method that scales to high-dimensional systems.
The large runtime of the gradient-based approach is mainly due to inefficient automatic differentiation of the MJX simulation, which remains the strongest limitation on gradient-based approaches at this point.
One potential improvement could be to implement implicit differentiation of the constraint solver in MJX, which could allow for much more efficient gradient computation.

\end{document}